\pdfoutput=1

\documentclass[11pt]{article}

\usepackage{EMNLP2022}

\usepackage{times}
\usepackage{latexsym}
\usepackage{graphicx}
\usepackage{algorithm2e}
\RestyleAlgo{ruled}
\usepackage{booktabs}
\usepackage{enumitem}
\usepackage{amsmath, amsfonts}
\usepackage{bm}
\usepackage{arydshln}
\usepackage{subfigure}
\usepackage{multirow}
\usepackage{tcolorbox}
\usepackage{amssymb}
\usepackage{pifont}
\newcommand{\cmark}{\ding{51}}%
\newcommand{\xmark}{\ding{55}}%

\newcommand{\mb}{\mathbf}

\newcommand{\bmu}{\bm{\mu}}
\newcommand{\bpi}{\bm{\pi}}
\newcommand{\bSigma}{\bm{\Sigma}}
\newcommand{\bLambda}{\bm{\Lambda}}

\newcommand{\calN}{\mathcal{N}}
\newcommand{\btheta}{\bm{\theta}}

\newcommand{\sslarge}{\textsubscript{{\fontfamily{qcr}\selectfont{large}}}}

\usepackage[T1]{fontenc}

\usepackage[utf8]{inputenc}

\usepackage{microtype}

\usepackage{inconsolata}

\title{Beyond Prompting: Making Pre-trained Language Models \\ Better Zero-shot Learners by Clustering Representations}

\author{Yu Fei$^1$, Zhao Meng$^{*1}$, Ping Nie$^{*2}$, Roger Wattenhofer$^1$, Mrinmaya Sachan$^1$\\ 
    $^1$ETH Zurich, $^2$Peking University\\ 
    \texttt{\{yu.fei, mrinmaya.sachan\}@inf.ethz.ch,} \\ 
    \texttt{\{zhmeng, wattenhofer\}@ethz.ch, ping.nie@pku.edu.cn}}

\begin{document}
\maketitle
\def\thefootnote{*}\footnotetext{Equal contribution.}\def\thefootnote{\arabic{footnote}}
\begin{abstract}

Recent work has demonstrated that pre-trained language models (PLMs) are zero-shot learners. However, most existing zero-shot methods involve heavy human engineering or complicated self-training pipelines, hindering their application to new situations. In this work, we show that zero-shot text classification can be improved simply by clustering texts in the embedding spaces of PLMs. Specifically, we fit the unlabeled texts with a Bayesian Gaussian Mixture Model after initializing cluster positions and shapes using class names. Despite its simplicity, this approach achieves superior or comparable performance on both topic and sentiment classification datasets and outperforms prior works significantly on unbalanced datasets. We further explore the applicability of our clustering approach by evaluating it on 14 datasets with more diverse topics, text lengths, and numbers of classes. Our approach achieves an average of 20\% absolute improvement over prompt-based zero-shot learning. Finally, we compare different PLM embedding spaces and find that texts are well-clustered by topics even if the PLM is not explicitly pre-trained to generate meaningful sentence embeddings. This work indicates that PLM embeddings can categorize texts without task-specific fine-tuning, thus providing a new way to analyze and utilize their knowledge and zero-shot learning ability\footnote{Code and datasets available at: \url{https://github.com/fywalter/simptc}}. 

\end{abstract}

\section{Introduction}
Recent developments in large pre-trained language models (PLMs) \citep{devlin-etal-2019-bert, liu2019roberta, JMLR:v21:20-074} open up the possibility of classifying texts without massive in-task data annotation. Such a zero-shot setting is receiving increasing attention as it is a good way to evaluate the generalizability of knowledge in PLMs. Currently, most existing methods either utilize keywords for self-training \citep{chang2008importance, meng2018weakly, wang-etal-2021-x} or reformulate the classification task into a cloze task using prompts \citep{NEURIPS2020_1457c0d6, schick-schutze-2021-exploiting, gao-etal-2021-making}. Keyword-based methods usually train multiple modules sequentially \citep{meng-etal-2020-text}, while prompting methods depend heavily on human engineering \citep{liu2021pre} or external knowledge \citep{hu2021knowledgeable}. Such task-specific training or engineering is inefficient and usually does not generalize well to new applications.

In this work, we show that we can better elicit the zero-shot text classification abilities of PLMs simply by clustering texts in their embedding spaces. We draw inspiration from recent findings \citep{aharoni-goldberg-2020-unsupervised} that texts in the same domain (e.g., legal or medical texts) tend to be clustered together in the PLM embedding spaces. This indicates that PLMs already have the knowledge to distinguish texts with different meanings. Following this idea, we propose \texttt{SimPTC}: A \textbf{Sim}ple \textbf{P}robabilistic \textbf{T}ext \textbf{C}lassification framework building upon state-of-the-art sentence embeddings SimCSE \citep{gao-etal-2021-simcse}. Given an unlabeled dataset and the corresponding class names, \texttt{SimPTC} models the texts in each class with a Gaussian distribution and fits the text embeddings with a Bayesian Gaussian Mixture Model (BGMM). To initialize the clusters, we first use the class names to generate class-related anchor sentences. Then the initial cluster assignment of a text is determined according to its similarity to the class anchors in the embedding space.

\par Despite the simplicity of \texttt{SimPTC}, it achieves state-of-the-art performance while avoiding many previously mentioned drawbacks of existing methods: 1) Without self-training of the PLM, \texttt{SimPTC} achieves superior or comparable performance on both topic and sentiment classification datasets; 2) Unlike prompt-based methods, \texttt{SimPTC} works well without human engineering or access to external knowledge; 3) \texttt{SimPTC} outperforms previous methods when the dataset is unbalanced. Finally, once we obtain the sentence embeddings, we no longer use the PLM, and \texttt{SimPTC} clusters the embeddings in a fixed dimensional space. Thus, one can easily apply \texttt{SimPTC} to new and large datasets. 

\par To explore the applications and limitations of \texttt{SimPTC}, we compare it with prompt-based zero-shot learning \citep{schick-schutze-2021-exploiting} on 14 datasets with more diverse topics, text lengths, and numbers of classes. \texttt{SimPTC} gives consistently better performance with a 20\% absolute improvement in macro-F1 score on average. We find that \texttt{SimPTC} handles domain-specific rare class names and large class numbers better, while both the prompt-based method and \texttt{SimPTC} suffer when the class names are abstract concepts, e.g., subjective v.s. objective.

\par Finally, we analyze the embedding spaces of different PLMs using \texttt{SimPTC}. Surprisingly, although RoBERTa\sslarge~\citep{liu2019roberta} is not explicitly pre-trained to generate meaningful sentence embeddings, texts of the same topic are clustered with state-of-the-art zero-shot accuracy. A Larger PLM like T5 \citep{raffel2020exploring} is able to achieve better zero-shot results, even matching the fully supervised performance of BERT \citep{devlin-etal-2019-bert} on some datasets. On the other hand, SimCSE embeddings separate topics better, and texts of sub-topics can form sub-clusters. On some datasets, we can even observe a linear semantic structure.



To conclude, the strong performance of such a simple clustering-based algorithm suggests that the zero-shot learning ability of PLMs is still under-explored. With \texttt{SimPTC}, we provide a new starting point to utilize and analyze the implicit knowledge and zero-shot learning ability of PLMs.

\begin{figure*}[t]
\centering
\includegraphics[width=\textwidth]{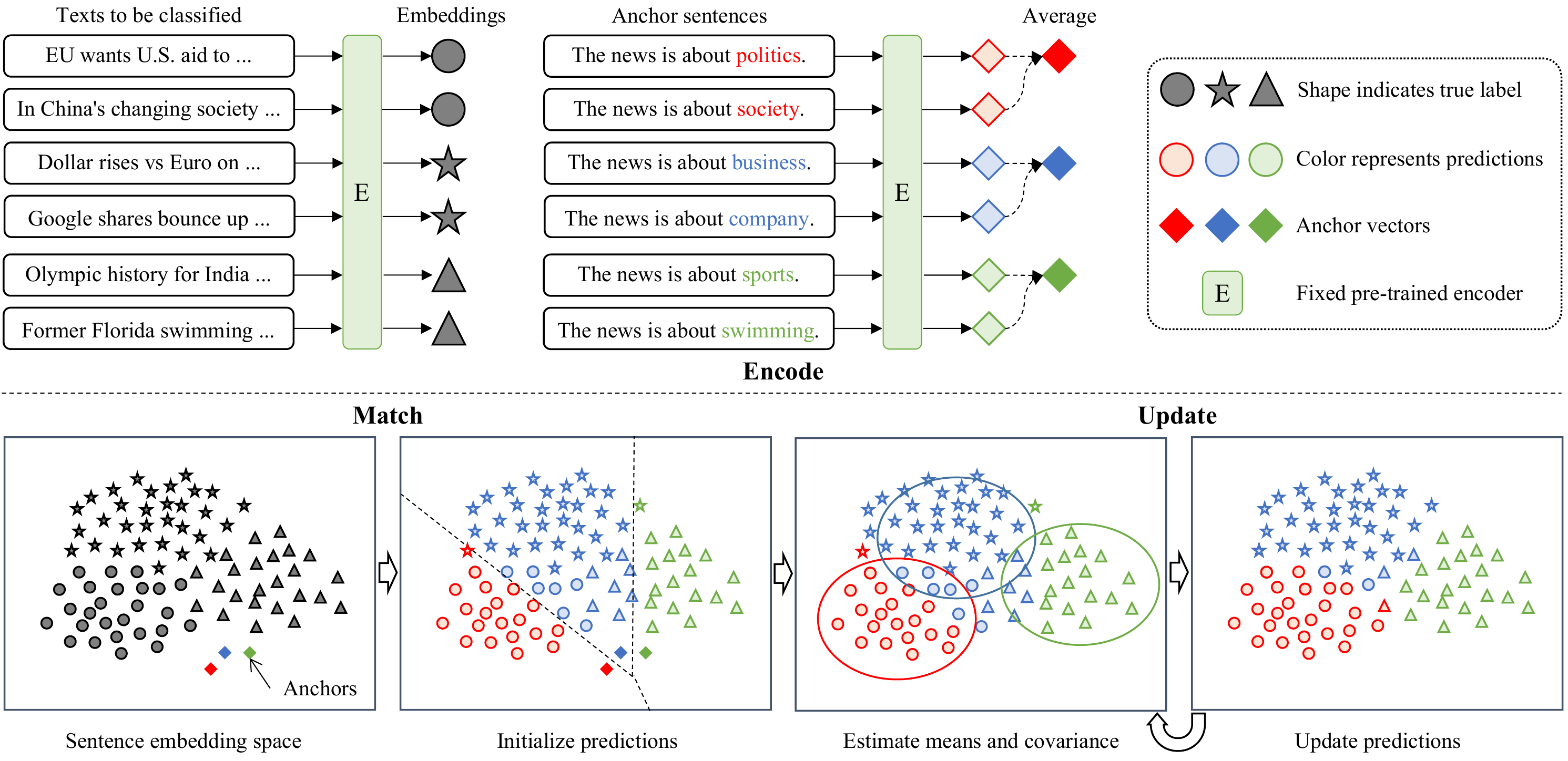}
\caption{An overview of \texttt{SimPTC}. \textbf{Top}: In the Encode step, all unlabeled texts and anchor sentences of each class are encoded using a PLM. Anchor sentences are constructed by combining a template with class names. The anchor sentence embeddings of the same class are averaged to get the final anchor vector. \textbf{Bottom left}: In the Match step, the initial cluster assignments are determined based on the cosine similarity between text embeddings and anchor vectors. \textbf{Bottom right}: In the update step, we fit the unlabeled data with a BGMM starting from the initial clusters.}
\label{fig:main}
\end{figure*}

\section{Related Work}
In this section, we review three types of zero-shot text classification approaches. Zero-shot text classification aims at classifying texts without any annotated data. This is also referred to as \emph{weakly-supervised} text classification as it can use various weak supervision signals, such as the names or descriptions of the classes, to make predictions. 
\paragraph{Keyword-driven methods} The most common supervision signal is keywords \citep{chang2008importance, mekala-shang-2020-contextualized}. \citet{meng2018weakly, meng-etal-2020-text} use iterative self-training on unlabeled in-task data to refine the model or keyword sets. \citet{wang-etal-2021-x} learn document representations that align with the classes. \citet{zhang-etal-2021-weakly} build a keyword graph to take the connections between keywords into account. Unlike these approaches, \texttt{SimPTC} contains no model training or keyword refinement process and depends solely on the sentence embedding spaces of PLMs. 

\paragraph{Clustering-based methods} Early clustering-based methods work with discrete text representations such as TF-IDF \citep{zeng2003cbc} or bag-of-words \citep{kyriakopoulou2006text}. Recently, ULR \citep{chu-etal-2021-unsupervised} has explored clustering-based text classification with contextualized sentence embeddings. However, ULR requires fine-tuning the PLM on extra task-related data and uses a heuristic regularization. The K-Means-based approach also places a strong spherical assumption on the cluster shapes. In this work, we show that neither the task-relevant pre-training nor the heuristic designs are necessary. The original embedding spaces of PLMs are sufficient to give strong results with a more flexible clustering algorithm. Nevertheless, it is possible to utilize unsupervised learning to further improve the clustering quality of text representations like in  \citet{gupta2022deep} and \citet{zhang2021supporting}. We leave this as a future direction.

\paragraph{Prompt-based methods}

Prompt-based methods perform zero-shot learning in a natural way by mimicking human behaviors when solving NLP tasks~\citep{NEURIPS2020_1457c0d6}. Many existing works on prompts focus on text classification, where a template is used to transform the classification task into a cloze task, and a verbalizer maps the predicted words into classification labels~\citep{schick-schutze-2021-exploiting}. With carefully designed templates and verbalizers, prompt-based methods can perform comparably to supervised methods in text classification.  
Various methods have been explored for designing templates \citep{gao-etal-2021-making, qin-eisner-2021-learning} and verbalizers \citep{cui2022prototypical}. Other researchers leverage external knowledge. \citet{hu2021knowledgeable} expand label names with knowledge bases, and \citet{chen2022adaprompt} re-train PLMs by adaptively retrieving extra data. 

\texttt{SimPTC} shares the idea of utilizing natural language templates and class names. Nevertheless, instead of reformulating the classification task, \texttt{SimPTC} uses natural language templates and class names to construct class-related texts, which are used to compute initial cluster positions and shapes for the subsequent probabilistic clustering step. 

\section{SimPTC}\label{sec:simptc}

As illustrated in Figure \ref{fig:main}, \texttt{SimPTC} formalizes a zero-shot text classification task into a clustering problem and solves it in three steps: Encode, Match, and Update. We start by modeling each class with a Gaussian cluster in the embedding space. 
Next, the Encode step and Match step provide a coarse initialization of the cluster means and covariances using the class names. Finally, starting from the initialization, we fit the unlabeled data with a BGMM. We elaborate on the three steps of \texttt{SimPTC} below.

\subsection{Encode}\label{sec:encode}
The first step of \texttt{SimPTC} is to construct class anchor sentences by filling the class names expanded based on external knowledge bases into natural language templates. Then we encode both the unlabeled texts and the class anchor sentences into the PLM embedding space (Figure \ref{fig:main} top).  

\paragraph{Expanding class names}
To make the anchor sentences more class-indicative and less dependent on the exact textual forms of the class names, we expand the class names using external knowledge bases. Specifically, we use ConceptNet Numberbatch \citep{speer2017conceptnet}, a set of word embeddings with semi-structured, common sense knowledge from ConceptNet \citep{speer2017conceptnet} combining word2vec \citep{mikolov2013efficient} and GloVe \citep{pennington-etal-2014-glove}. To extract $M$ related words given a class name $s_i$, we simply choose the words whose embeddings have top-$M$ largest inner products with the embedding of $s_i$: 
\begin{equation*}
    S_i = \mathop{\text{top-}M}_{x\in \mathcal{V}}(\mb x^\top \mb s_i),
\end{equation*}
where $S_i$ is the expanded class name set of $s_i$; $\mathcal{V}$ is the vocabulary; bold font denotes word embeddings. Words that appeared in multiple $S_i$'s are deleted. If $m>1$ class names are given for one class, for each name we extract $M/m$ words. See Appendix \ref{sec:ap_class_names} for extracted word examples.

\paragraph{Constructing anchor sentences}
We take the idea of using natural language templates from prompt-based methods \citep{schick-schutze-2021-exploiting} to construct anchor sentences. A \emph{template} is a piece of text containing one or multiple special tokens to be filled in, such as ``$\text{The text is about } \langle mask \rangle.$''. By replacing the $\langle mask \rangle$ token with the expanded class names $s_i\in S_i$, we get a set of class-related sentences. Unlike prompt-based methods, we allow class names with multiple tokens. 
The anchor embeddings of the same class are averaged to give the final anchor vector (Figure \ref{fig:main} top middle). 

\begin{figure}[t]
\centering
\includegraphics[width=0.6\columnwidth]{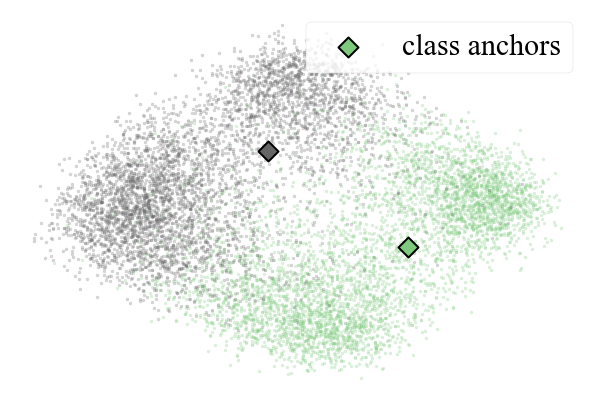}
\caption{2D PCA visualization of Amazon dataset in the SimCSE embedding space. \textbf{The texts of different classes are well-clustered, and the class anchors from the Encode step reflect the relative positions of text clusters. (Figure \ref{fig:main} bottom left).}}
\label{fig:dbpedia_pca}
\end{figure}
\subsection{Match}\label{sec:match}
Let $\{\mb x_i, \mb x_2, \dots, \mb x_N\}$ be the embeddings of a set of unlabeled texts of size $N$, and $\{\mb a_1, \mb a_2,\dots, \mb a_K\}$ be the averaged anchor vectors for $K$ classes. The pseudo-label $\hat{y}_i$ of $\mb x_i$ are determined by:
\begin{equation}\label{eq:match}
    \hat{y}_i = \mathop{\arg\max}_{j\in [K]} \text{cos-sim}(\mb x_i, \mb a_j).
\end{equation}
$\{\hat{y}_i\}$ are then used to compute the initial cluster means and covariances. We call this pseudo-label-generating process \texttt{Encode\&Match} \texttt{(E\&M)}. 



Figure~\ref{fig:dbpedia_pca} illustrates the encoded anchor vectors $\mb a_i$'s and example vectors $\mb x_i$'s after performing PCA. The anchor vectors $\mb a_i$'s indeed reflect the relative positions of the clusters. To provide more insights, we conduct a pilot experiment on AG's News \citep{NIPS2015_250cf8b5} dataset . We show the zero-shot performance of \texttt{E\&M} and \texttt{Vanilla Prompting}, the vanilla prompt-based zero-shot text classification method used in \citet{schick-schutze-2021-exploiting}, in Table \ref{tab:ag_compare}\footnote{\texttt{Vanilla Prompting} results are based on BERT\sslarge, and \texttt{E\&M} uses SimCSE supervised BERT\sslarge}. For a fair comparison, we use the original class names directly to construct anchor sentences for \texttt{E\&M}. \texttt{E\&M} provides a competitive initialization and is more stable across different choices of natural language templates.

\begin{table}
\centering
\small
\begin{tabular}{lc}
\toprule
\textbf{Template} & \textbf{Acc} \\ 
	\midrule
	\multicolumn{2}{c}{\texttt{Vanilla Prompting}}\\
	\midrule
 $\langle text\rangle$ A $\langle mask\rangle$ news .                                     & 31.5\\
 $\langle text\rangle$ [class: $\langle mask\rangle$]                                  & 70.3\\
 $\langle text\rangle$ This text is about $\langle mask\rangle$ .                         & 68.7\\
 	\midrule
 \multicolumn{2}{c}{\texttt{Encode\&Match}}\\
 	\midrule
  A $\langle mask\rangle$ news.                                     & \textbf{78.9}\\
 \ [class: $\langle mask\rangle$]                                & \textbf{76.8}\\
  This text is about $\langle mask\rangle$.                         & \textbf{78.2}\\
\bottomrule
\end{tabular}
\caption{Accuracy of \texttt{Vanilla Prompting} and \texttt{Encode\&Match} with different templates on AG's News test set. \textbf{\texttt{Encode\&Match} depends less on the choice of template and gives better performance.}}
\label{tab:ag_compare}
\end{table}
\subsection{Update}
\paragraph{Model classes with Gaussian clusters} To capture the position and shape characteristics of text clusters, we model the texts of the same class with a Gaussian in the embedding space and define a Gaussian Mixture Model (GMM). Then the likelihood of the dataset is given by
\begin{equation*}
   p(\mb X|\btheta)=\prod_{n=1}^N \sum_{i=1}^{K} \pi_i \mathcal{N}(\mb x_n|\bmu_i,\bSigma_i),
\end{equation*}
where $\btheta=(\bpi,\bmu,\bSigma)$ denotes the model parameters; $\bpi=\{\pi_1, \dots,\pi_K\}$, $\bmu=\{\bmu_1,\dots,\bmu_K\}$, and $\bSigma=\{\bSigma_1,\dots,\bSigma_K\}$ are the priors, means, and covariances of each component respectively. We can further require all components to share the same covariance matrix to add extra regularization when the data is sparse, or we have additional prior knowledge that the clusters have similar shapes.
\paragraph{Variational update} Clustering in a high dimensional space can be challenging, for instance, when the data is limited, or the initialization is poor. One simple solution is to inject prior knowledge, such as assuming a uniform prior on the classes as in several prompt-based methods \citep{pmlr-v139-zhao21c, hu2021knowledgeable}. However, debiasing model explicitly can be harmful when the prior is incorrect. To balance injecting prior knowledge and fitting the data, we turn to the Bayesian approaches and introduce prior distributions on model parameters. We choose a Dirichlet distribution as the prior for mixture weights $\bpi$ to favor balanced weights:
\begin{equation*}
    p(\bpi) =~\text{Dir}(\bpi|\alpha_0)=C(\alpha_0)\prod_k \pi_k^{\alpha_0-1}
\end{equation*}
where $C(\alpha_0)$ is a normalizing constant, and $\alpha_0$ can be interpreted as the prior number of observations associated with each class. We simply choose $\alpha_0=N/K$. For the means and covariances, we choose a non-informative Gaussian-Wishart prior (see Appendix \ref{sec:ap_update} for details). Then we update the model with the standard variational optimization \citep{bishop2006pattern}. As BGMM is guaranteed to converge \citep{boyd2004convex}, we stop updating when the model predictions stop changing or the maximum number of iterations is reached.

The overall \texttt{SimPTC} algorithm is summarized in Algorithm \ref{alg:SimPTC}. \textbf{Note that, in general, we can replace \texttt{Encode\&Match} with any initialization method and Bayesian GMM with any clustering algorithm} (see discussion in \S\ref{sec:ablation}).

\begin{algorithm}
\SetKwInput{Input}{Input}
\SetKwInput{Output}{Output}
\caption{SimPTC}\label{alg:SimPTC}
\Input{unlabeled texts $U$; test texts $U^{test}$; class names $S$; sentence encoder $E$; max iteration $T$}
\Output{The prediction of $U^{test}$}
$\mb X\leftarrow E(U)$;\\
$\mb X^{test}\leftarrow E(U^{test})$;\\
$\{\hat{y}_i\}\leftarrow$ \texttt{Encode\&Match} (\S\ref{sec:encode} and \S\ref{sec:match});\\ 
$M\leftarrow$ BayesianGMM(\\
\quad initial predictions $\leftarrow \{\hat{y}_i\}$,\\
\quad weight prior $\alpha_0\leftarrow |U|/|S|,$\\
\quad mean \& cov prior $\leftarrow$ Eq. \eqref{eq:prior} in App. \ref{sec:ap_update},\\
\quad max iter $\leftarrow T$,\\
);\\
Fit $M$ with $\mb X$;\\
$\{y^{test}_i\}\leftarrow$ prediction of $M$ on $\mb X^{test}$;\\
\textbf{Return} $\{y^{test}_i\}$
\end{algorithm}

\begin{table*}
\centering
\small
\begin{tabular}{llllll}
\toprule
\textbf{Method} & \textbf{AG's News} & \textbf{DBPedia} & \textbf{Yahoo} & \textbf{Amazon} & \textbf{IMDb}\\
\midrule
\texttt{ULR}  \citep{chu-etal-2021-unsupervised}  & $80.1$ & $79.8$ & $59.6$ & $92.6$ & $82.4$ \\
\texttt{LOTCLass}$\dagger$  \citep{meng-etal-2020-text}  & $ 86.4$ & $91.1 $ & fail & $91.6$ & $86.5$ \\
\texttt{Vanilla Prompting}  & $ 72.1\pm 10.4 $ & $ 80.9\pm 2.3$ & $40.4\pm 3.1 $ & $79.7\pm 10.8 $ & $81.5\pm 4.1$ \\
\texttt{KPT}$\dagger$  \citep{hu2021knowledgeable}  & $84.8 \pm 1.2 $ & $ 82.2 \pm 5.4$ & $61.6 \pm 2.2 $ & $92.8 \pm 1.2 $ & $\bf 91.6 \pm 2.7 $ \\
\midrule
\texttt{Encode\&Match (E\&M)} & $78.2 \pm 0.3 $ & $ 74.4 \pm 1.6$ & $58.3 \pm 0.1 $ & $91.2 \pm 0.1 $ & $85.6 \pm 0.4 $ \\
\texttt{SimPTC} & $\bf 86.9 \pm 0.3 $ & $\bf 93.2 \pm 1.0$ & $\bf 63.9 \pm 0.1 $ & $\bf 93.9 \pm 0.0 $ & $\bf 91.0 \pm 0.0 $ \\
\midrule
\quad-class name expansion & $87.6 \pm 0.5 $ & $ 92.9 \pm 0.1$ & $63.7 \pm 0.1 $ & $93.9 \pm 0.0 $ & $91.0 \pm 0.0 $ \\
\quad\quad-manual templates  & $86.5$ & $ 93.3$ & $62.9$ & $93.9$ & $91.1$ \\
\bottomrule
\end{tabular}
\caption{\label{tab:main}
Zero-shot test accuracy on five benchmark datasets. $\dagger$: We use the number reported in the original papers. Indentation means the configuration is modified based on the up-level indentation. The keyword-extracting module of \texttt{LOTCLass} fails on Yahoo.
}
\end{table*}
\section{Experiments}\label{sec:experiment}
We conduct extensive experiments to understand \texttt{SimPTC}. We compare \texttt{SimPTC} with state-of-the-art zero-shot text classification methods in \S\ref{sec:sota}, study the effect of its components in \S\ref{sec:ablation}, and explore its applications and limitations on a wide range of tasks in \S\ref{sec:tc14}. For all experiments, we use the SimCSE supervised RoBERTa\sslarge~embeddings, which are in $\mathbb{R}^{1024}$ and trained using NLI datasets via contrastive learning starting from the original RoBERTa\sslarge~ model. We discuss and analyze other PLMs, such as T5 in \S\ref{sec:encoder}.
\subsection{Comparison with State-of-the-art}\label{sec:sota}
We evaluate the zero-shot text classification performance of \texttt{SimPTC} on five benchmark datasets.
\paragraph{Datasets} We use three topic datasets: AG's News \citep{NIPS2015_250cf8b5}, DBpedia \citep{lehmann2015dbpedia}, and Yahoo \citep{NIPS2015_250cf8b5}, and two sentiment datasets: IMDb \citep{maas-etal-2011-learning} and Amazon \citep{10.1145/2507157.2507163}. The full dataset statistics can be found in Appendix \ref{sec:ap_dataset}.
\paragraph{Implementations}
Following \citet{hu2021knowledgeable}, we manually design four templates (Appendix \ref{sec:ap_templates}) for every dataset. The number of extracted class-related words for each class is $1000$. We fit the BGMM with both the unlabeled train and test data. For topic datasets, each Gaussian has its individual covariance. For sentiment datasets, all Gaussians share the same covariance to provide extra regularization as the data is relatively sparse. The maximum iterations are set empirically based on the size of unlabeled data. See Appendix \ref{sec:ap_dataset} for details.
\paragraph{Baselines} We compare \texttt{SimPTC} with the following methods. \texttt{Vanilla Prompting} is the vanilla prompt-based zero-shot text classification without self-training used in \citet{schick-schutze-2021-exploiting}. We use the original class names and templates designed by \citet{hu2021knowledgeable} for predicting. \texttt{ULR} \citep{chu-etal-2021-unsupervised} performs zero-shot text classification by clustering data using K-Means with a heuristic regularization. Since \texttt{ULR} originally uses an encoder pre-trained with extra in-domain data, we evaluate \texttt{ULR} with the same embeddings used by \texttt{SimPTC}. \texttt{LOTCLass} \citep{meng-etal-2020-text} is a state-of-the-art keyword-based method that involves training multiple models with multiple tasks sequentially. \texttt{KPT} \citep{hu2021knowledgeable} is the state-of-the-art prompt-based method that utilizes external knowledge bases and contextualized calibration to produce stable zero-shot predictions. 
\paragraph{Experimental Design}
We conduct experiments to evaluate the following three claims:

\begin{figure*}[t]
\centering
\subfigure[Micro-F1 score]{\label{fig:unbalanced_micro}\includegraphics[width=0.37\textwidth]{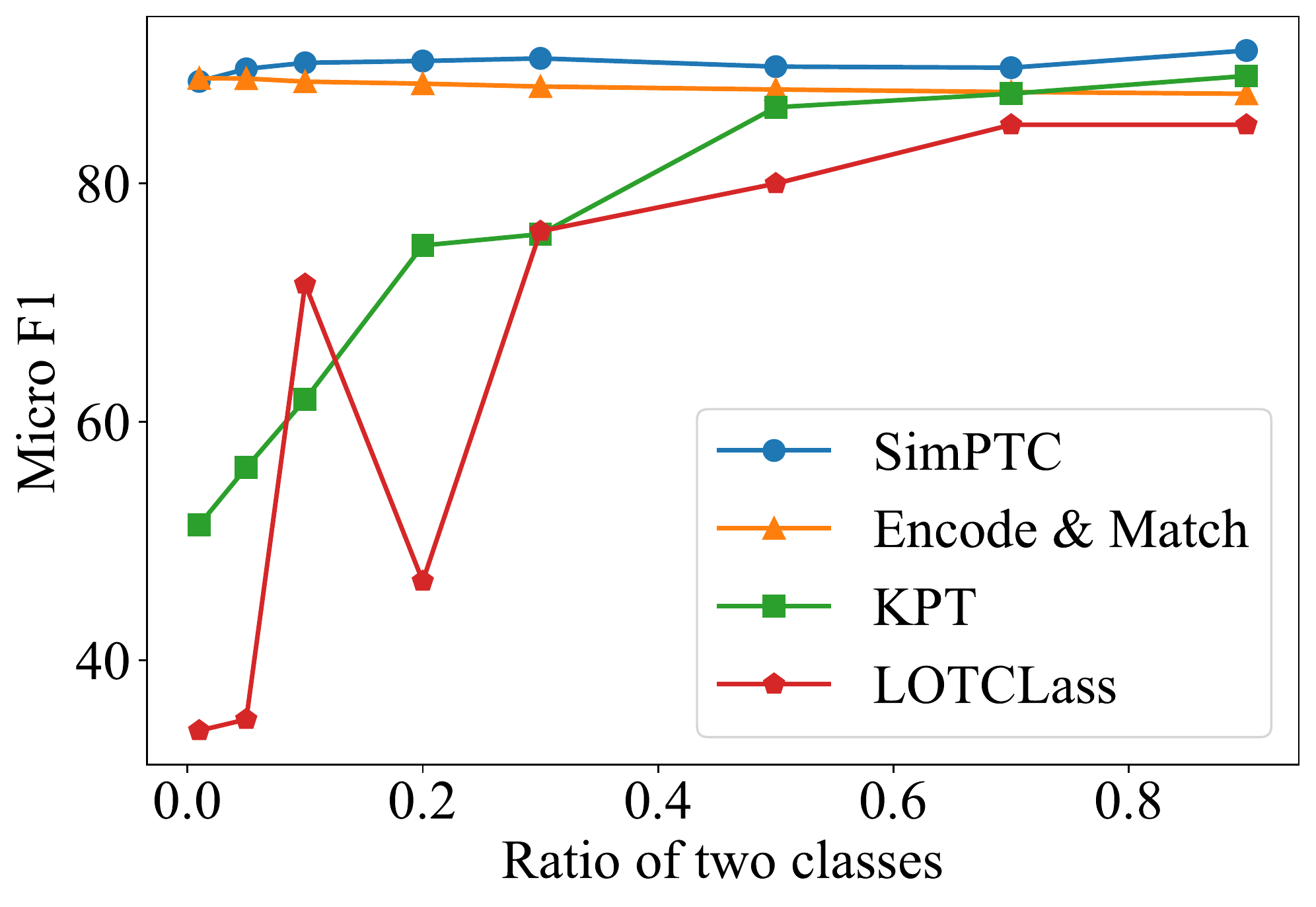}} 
\subfigure[Macro-F1 score]{\label{fig:unbalanced_macro}\includegraphics[width=0.37\textwidth]{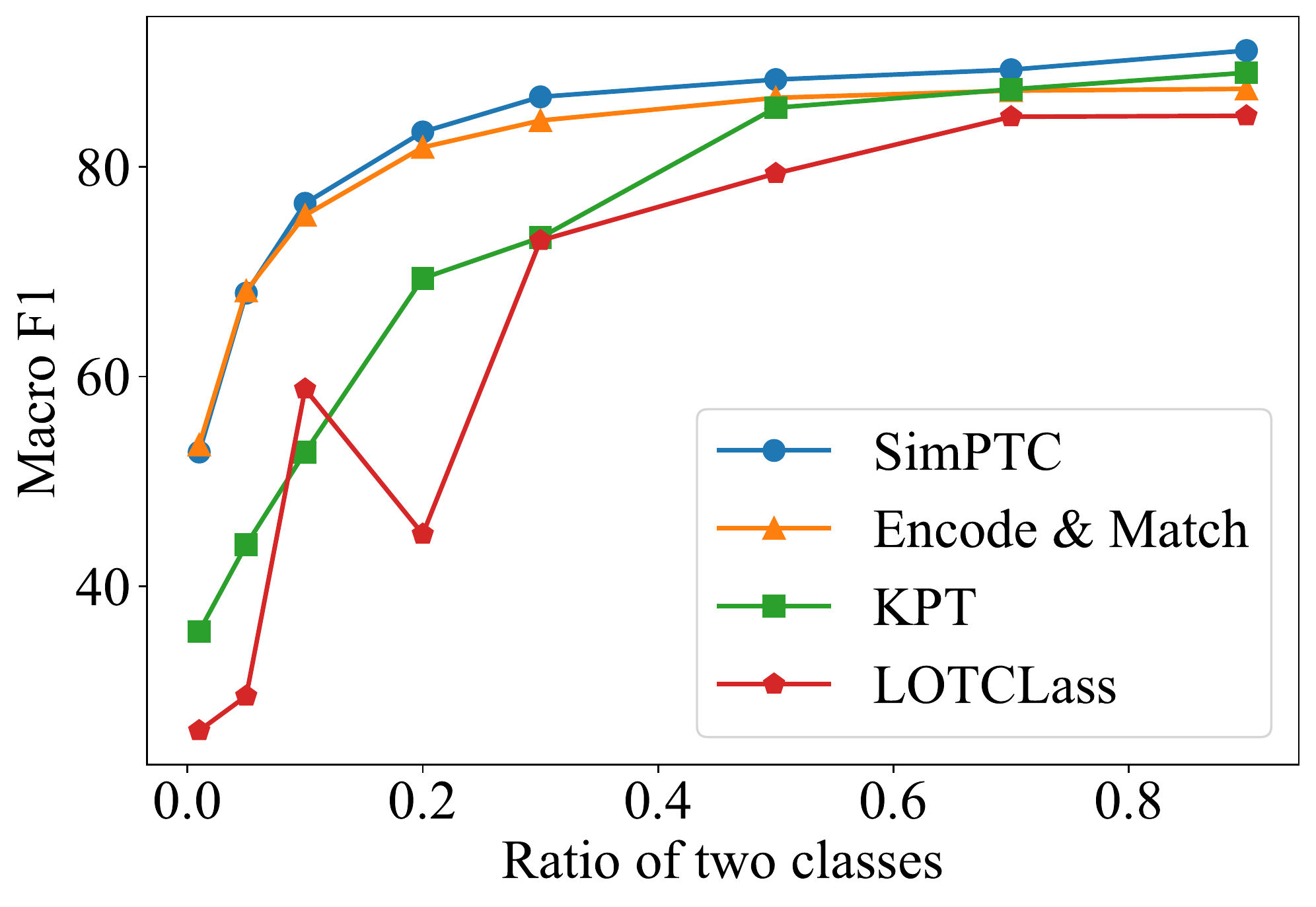}} 
\caption{The micro-F1 and macro-F1 score plots of different methods on unbalanced IMDb datasets with different class ratios. \textbf{When the dataset is unbalanced, \texttt{SimPTC} consistently performs better, and the gain is substantial in extreme cases.}}
\label{fig:unbalanced}
\end{figure*}

\paragraph{C1: \texttt{SimPTC} achieves superior or comparable performance on both topic and sentiment datasets.} Table \ref{tab:main} reports the accuracy on the test sets. We report the average scores with standard deviations for methods using multiple natural language templates. Without fine-tuning the PLM, \texttt{SimPTC} presents superior or comparable performance on all datasets. On IMDb, \texttt{KPT} gets slightly better results ($91.6$ v.s. $91.0$) but has a much larger standard deviation ($2.7$ v.s. $0$). Moreover, \texttt{KPT} improperly poses a balanced dataset assumption (Appendix \ref{sec:ap_kpt}), which hurts model performance when the dataset is unbalanced (see \textbf{C3}). 

\paragraph{C2: \texttt{SimPTC} gives stable predictions across different templates. }Compared to \texttt{Vanilla Prompting}, \texttt{E\&M} gives a better or comparable performance on all datasets with much lower standard deviations across different natural language templates (Table \ref{tab:main}). The observation holds even when we compare \texttt{E\&M} with the prompt-based method enhanced with external knowledge (\texttt{KPT}). \texttt{SimPTC} further reduces the standard deviations and improves performance. 

\paragraph{C3: \texttt{SimPTC} consistently outperforms prior work when the classes in the dataset are unbalanced. }Currently, most benchmark datasets are balanced. Overfitting to this balanced bias reduces the generalizability of the method. To illustrate this problem, we conduct the following experiment on IMDb. We keep the texts of one class with a ratio varying from $0.01$ to $0.9$ to generate different unbalanced settings, and we compare \texttt{SimPTC} with \texttt{KPT} and \texttt{LOTCLass}. \texttt{KPT} injects a balanced dataset assumption directly into its design (Appendix \ref{sec:ap_kpt}), and \texttt{LOTCLass} is a self-training keyword-based method without an explicit balanced assumption. As shown in Figure \ref{fig:unbalanced}, the performance of \texttt{KPT} and \texttt{LOTClass} drops significantly as the dataset becomes more unbalanced, whereas \texttt{SimPTC} achieves consistently better performance. As the class ratio approaches zero, the micro-F1 score of \texttt{KPT} goes to $50$ since the balanced prior forces the model to make a balance prediction. Although \texttt{LOTClass} is purely data-driven, the data imbalance still dramatically affects its self-training process. On the other hand, \texttt{E\&M} provides a strong starting point for \texttt{SimPTC}, and \texttt{SimPTC} further improves its performance.

We discuss the convergence of \texttt{SimPTC}, the effect of unlabeled dataset size, and sharing covariance matrix in Appendix \ref{sec:ap_converge}, \ref{sec:ap_dataset_size} and \ref{sec:ap_covariance} respectively.

\subsection{Ablations}\label{sec:ablation}
We try to understand what contributes to the competitive performance of \texttt{SimPTC} by studying the importance of 1) the choice of natural language template and class names, 2) the initialization method, and 3) the clustering algorithm.
\subsubsection{Templates and Class Names}
\paragraph{\texttt{SimPTC} gives state-of-the-art results even without carefully designed templates or class names extracted using external knowledge. }We first evaluate \texttt{SimPTC} using only the original class names for constructing class anchor sentences (-class name expansion in Table \ref{tab:main}). \texttt{SimPTC} still gives a comparable performance on all datasets. Then we further test \texttt{SimPTC} with the naive template ``$\langle mask \rangle$'' (-manual templates in Table \ref{tab:main}). The performance is again only slightly affected. Unlike prompt-based methods, which are sensitive to the quality of class names and templates, \texttt{SimPTC} gives strong performance even without external knowledge or human engineering.
\begin{table}
\centering
\small
\begin{tabular}{llllll}
\toprule
\textbf{Method}& \textbf{AG} & \textbf{DB} & \textbf{YH} & \textbf{AM} & \textbf{IM}\\
\midrule
\texttt{VP}         & $72.1$ & $80.9$ & $40.4$ & $79.7$ & $81.5$\\
\texttt{E\&M}          & $78.2$ & $74.4$ & $58.3$ & $91.2$ & $85.6$\\
\midrule
\texttt{SimPTC}+\texttt{VP}   & $86.7$ & $92.7$ & $63.4$ & $\textbf{93.9}$ & $\textbf{91.0}$\\
\texttt{SimPTC}+\texttt{E\&M}  & $\textbf{86.9}$ & $\textbf{93.2}$ & $\textbf{63.9}$ & $\textbf{93.9}$ & $\textbf{91.0}$\\
\bottomrule
\end{tabular}
\caption{\label{tab:initialization}
Comparison of different initialization methods. \textbf{\texttt{SimPTC} is fairly robust to the quality of initialization.
}}
\end{table}
\subsubsection{Initialization Method}
\paragraph{\texttt{SimPTC} is robust to the quality of initialization.} We use \texttt{E\&M} to initialize the clusters mainly because \texttt{E\&M} works directly with the text embeddings computed for later clustering, adding only minimal additional computations. In general, \texttt{SimPTC} works with any initialization method (see Algorithm \ref{alg:SimPTC}). As a comparison, we test using \texttt{Vanilla Prompting(VP)} as the initialization. We report the results averaged over four templates on five benchmarks in Table \ref{tab:initialization}. Although \texttt{VP} gives a slightly worse initialization performance, \texttt{SimPTC} achieves a similar performance after clustering, showing the robustness of \texttt{SimPTC} to the initialization method.

\begin{table}
\centering
\small
\begin{tabular}{llllll}
\toprule
\textbf{Clustering Algo.}& \textbf{AG} & \textbf{DB} & \textbf{YH} & \textbf{AM} & \textbf{IM}\\
\midrule
\#Class         & 4      &  14    &    10  &     2  &  2\\   
\midrule
K-Means         & $75.3$ & $90.5$ & $61.7$ & $92.1$ & $88.3$\\
GMM             & $76.4$ & $82.9$ & $51.6$ & $\textbf{93.9}$ & $89.4$\\
BGMM            & $\textbf{86.9}$ & $\textbf{93.2}$ & $\textbf{63.9}$ & $\textbf{93.9}$ & $\textbf{91.0}$\\
\bottomrule
\end{tabular}
\caption{\label{tab:clustering}
Comparison of different clustering algorithms. \textbf{BGMM outperforms K-Means, while GMM fails to work on many-class tasks like DBpedia and Yahoo}.
}
\end{table}

\subsubsection{Clustering Algorithm}
In this section, we aim to show what makes a good choice of clustering algorithm for \texttt{SimPTC} by comparing BGMM with K-Means and GMM.
 \paragraph{Modeling cluster shapes is beneficial.} As shown in Table \ref{tab:clustering}, BGMM outperforms K-Means on all five balanced benchmark datasets. This shows that putting a strong assumption on the cluster shapes like K-Means limits the clustering step's performance. Since the SimCSE embedding space is rather well-structured, we further test \texttt{SimPTC} + K-Means with the original RoBERTa\sslarge~embeddings. The performance on IMDb drops from 92.3 to 54.1, indicating that BGMM is a more robust choice for clustering PLM embeddings in general. 

 \paragraph{Adding prior on cluster weights helps on many-class tasks.} Following the previous observation, GMM outperforms K-Means on AG News, IMDb, and Amazon by allowing to model the cluster shapes using data. However, GMM fails on many-class tasks like DBpedia and Yahoo (Table \ref{tab:clustering}), showing the benefits of adding prior on cluster weights as extra regularization.
\begin{figure}[t]
\centering
\includegraphics[width=0.37\textwidth]{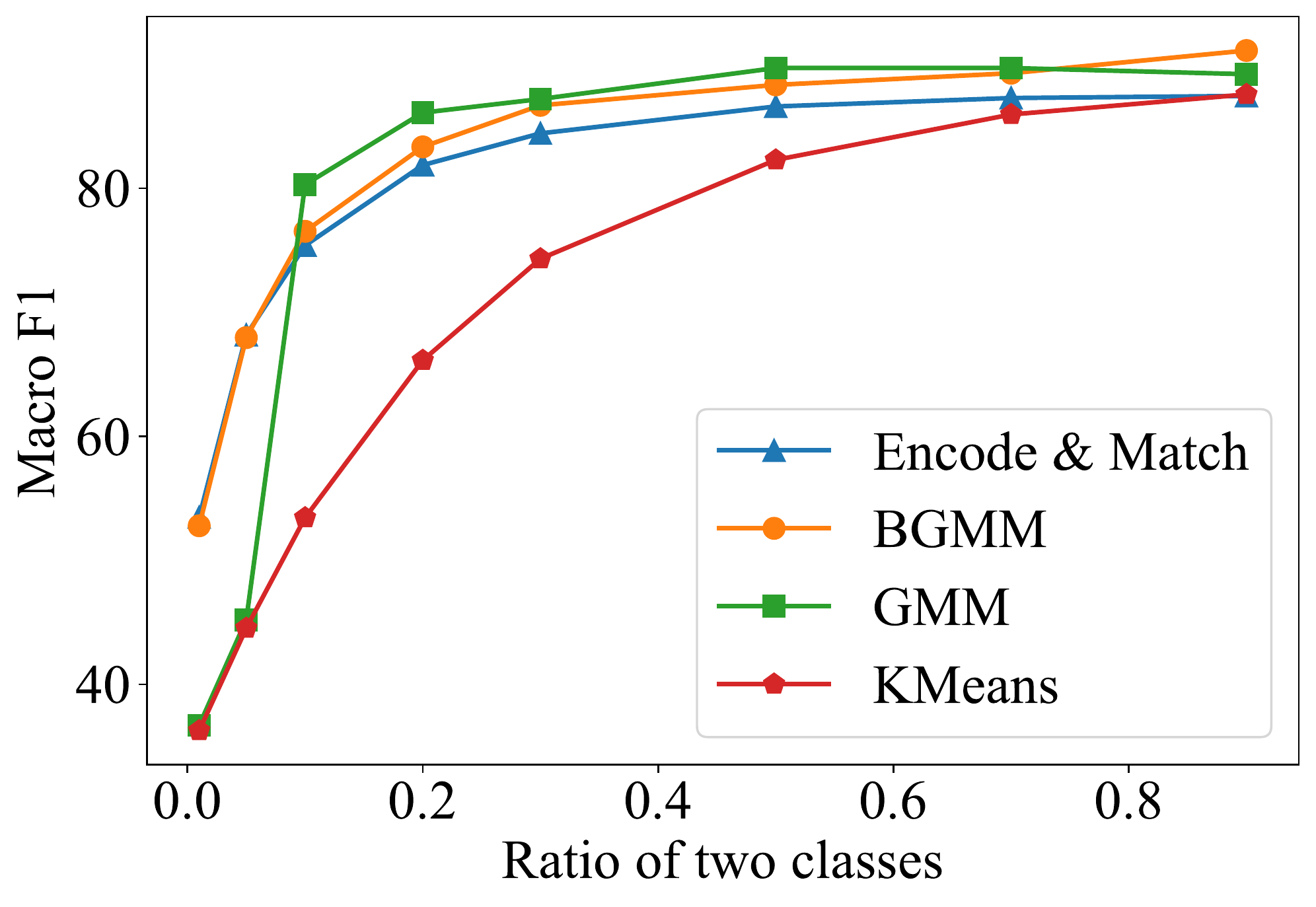}
\caption{The macro-F1 score plots of different clustering algorithms on unbalanced IMDb datasets with different class ratios. \textbf{K-Means cannot handle unbalanced datasets. BGMM and GMM perform better by allowing the cluster weights to adapt to the data, but GMM is less stable in extreme cases.}}
\label{fig:unbalanced}
\end{figure}
 \paragraph{Learnable cluster weights handle class imbalance.} The learnable mixing weights of BGMM (and GMM) model the proportion of classes and therefore handle unbalanced clusters. To test this, we again compare three clustering algorithms on IMDb dataset with different class ratios. Figure \ref{fig:unbalanced} shows that K-means fails completely when the dataset is unbalanced. BGMM and GMM perform better by allowing the cluster weights to adapt to the data, but GMM is less stable in extreme cases.
\begin{figure*}[t]
\centering
\includegraphics[width=\textwidth]{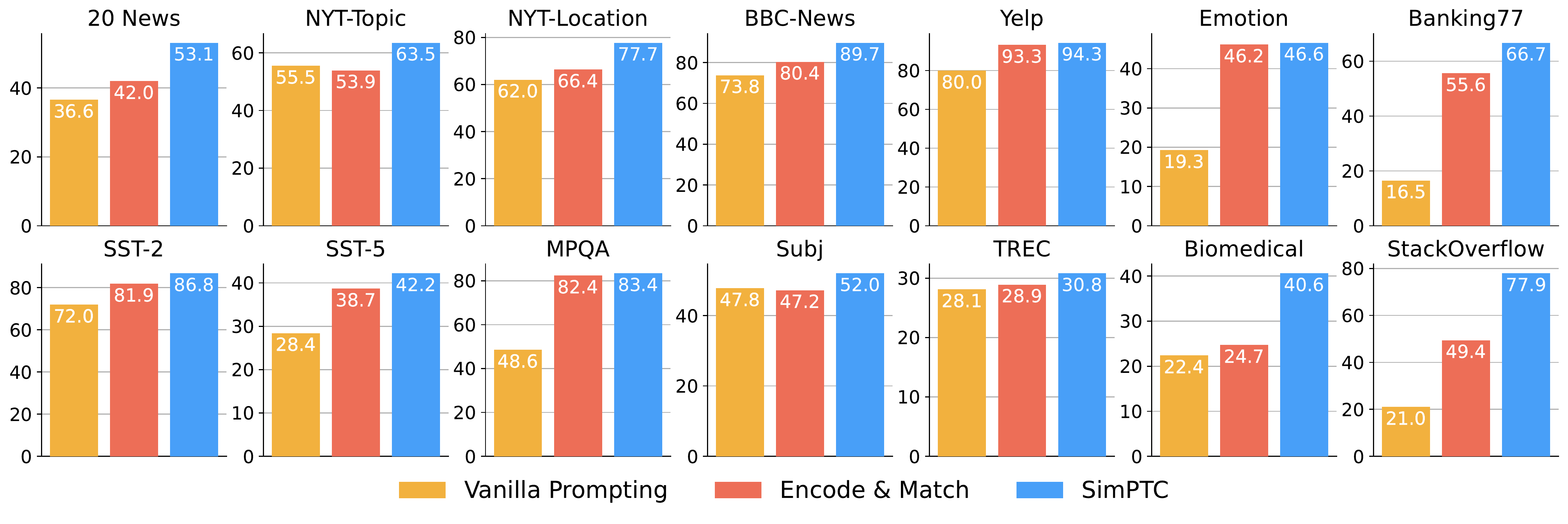}
\caption{Macro-F1 scores on TC14. \textbf{\texttt{SimPTC} outperforms \texttt{Vanilla Prompting} on all 14 datasets.}}
\label{fig:tc14}
\end{figure*}
\subsection{TC14 Datasets}\label{sec:tc14}
To further study the potential applications and limitations of \texttt{SimPTC}, we collect 14 publicly available text classification datasets with various topics, text lengths, and numbers of classes (Table \ref{tab:other_datasets_stat}). For simplicity, we refer to these datasets as TC14. For more dataset information, see Appendix \ref{sec:ap_tc14_info}. 
\paragraph{Setup} To simulate the most basic scenario, we evaluate \texttt{SimPTC} with the naive template ``$\langle mask \rangle$'' and the original class names without expansion. We choose \texttt{Vanilla Prompting} as the baseline since it is the most widely used zero-shot prompt-based method. For a fair comparison, we do not engineer templates or verbalizers and use the original class names with templates adopted from \citet{hu2021knowledgeable} (see Appendix \ref{sec:ap_tc14_implementation} for implementation details).

\begin{table}[t]
\centering
\small
\begin{tabular}{ccccc}
\toprule
 \textbf{Datasets} &\textbf{\# Texts} &\textbf{\# Cls.} &\textbf{Ave. Len.} & \textbf{Unb.}\\
 \midrule
20 News &		18391&	20&	186 &\cmark\\
NYT-Topic&		31997&	9&	783&\cmark\\
NYT-Location&		31997&	10&	783&\cmark\\
BBC News	&	2225&	5&	390&\cmark\\
Yelp		&38000&	2	&132&\xmark\\
Emotion		&20000&	6	&19&\cmark\\
Banking77	&	13083&	77&	12&\cmark\\
SST-2		&9613	&2&	19&\cmark\\
SST-5		&11855	&5&	19&\cmark\\
MPQA		&10606	&2&	3&\cmark\\
Subj		&10000	&2&	23&\xmark\\
TREC		&5952	&6&	10&\cmark\\
Biomedical	&	20000&	20&	13&\xmark\\
StackOverﬂow&		20000&	20&	8&\xmark\\
\bottomrule
\end{tabular}
\caption{\label{tab:other_datasets_stat}
TC14 datasets (Cls.: class, Unb.: unbalanced).
}
\end{table}
\paragraph{Results} We report the macro-F1 scores on TC14 in Figure \ref{fig:tc14} and put micro-F1 scores in Appendix \ref{sec:ap_tc14_results}. \texttt{E\&M} outperforms \texttt{Vanilla Prompting} on 12 out of 14 datasets. \texttt{SimPTC} further boosts the performance and gives a superior performance on all 14 datasets, showing the strong generalizability of our approach. \texttt{SimPTC} achieves the most gain when 1) the class names contain multiple tokens (e.g., Banking77); 2) the number of classes is large (e.g., StackOverflow); 3) the class names contain rare or domain-specific words (e.g., Biomedical). 
\begin{figure}[t]
\centering
\includegraphics[width=0.6\columnwidth]{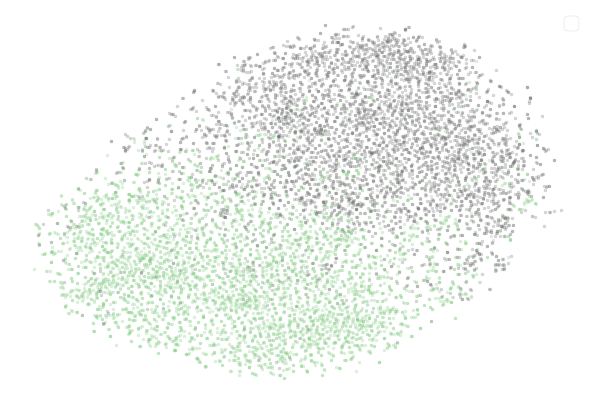}
\caption{2D t-SNE visualization of Subj dataset in the SimCSE embedding space. \textbf{Although \texttt{E\&M} cannot provide a meaningful initial prediction given the abstract class names: subjective and objective, the two classes are well separated in the embedding space.}}
\label{fig:tsne_subj}
\end{figure}
\begin{figure*}[t]
\centering
\subfigure[SimCSE-RoBERTa\sslarge.]{\includegraphics[width=0.55\columnwidth]{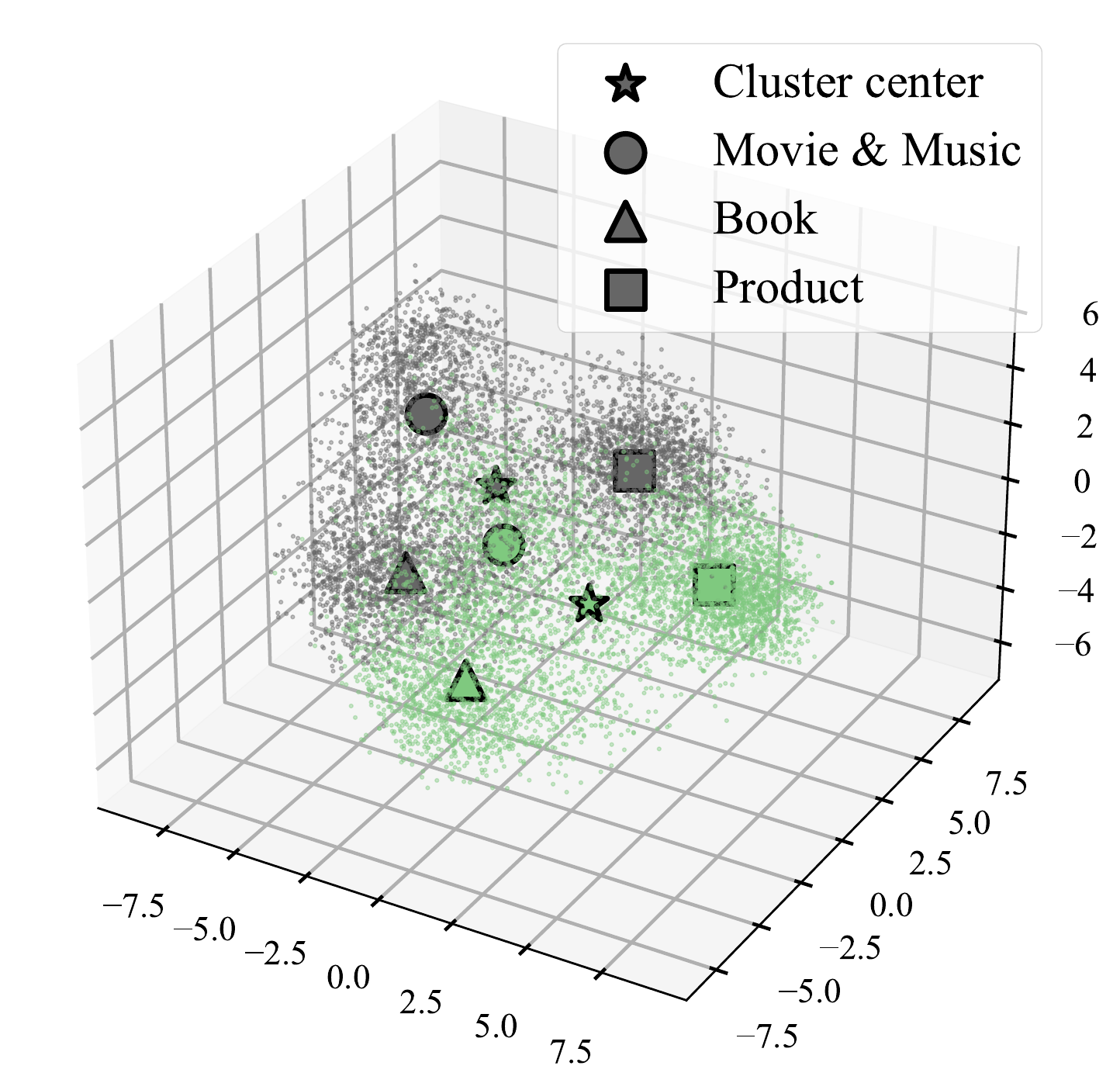}} 
\subfigure[RoBERTa\sslarge.]{\includegraphics[width=0.55\columnwidth]{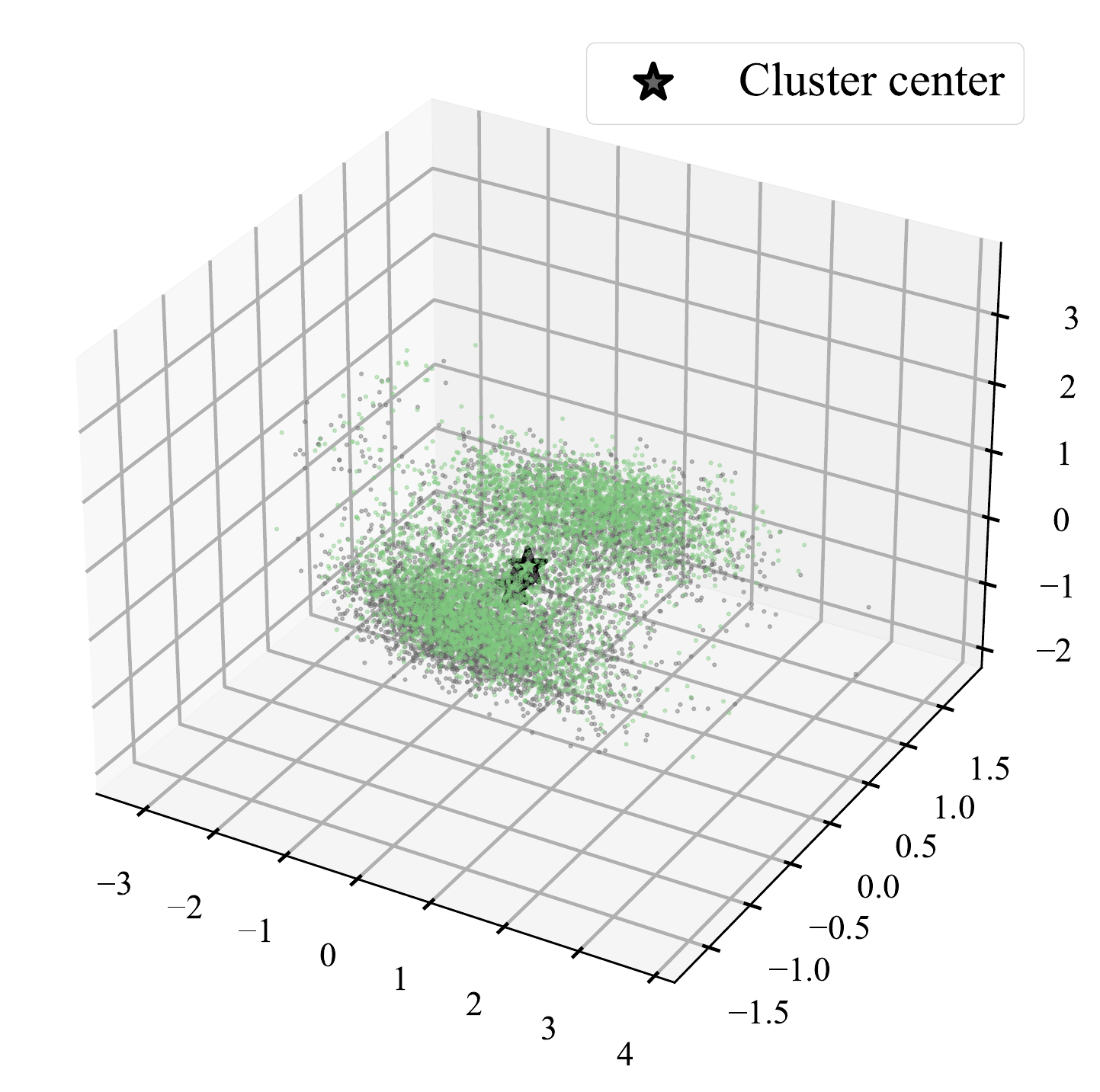}} 
\subfigure[Sentence-RoBERTa\sslarge.]{\includegraphics[width=0.55\columnwidth]{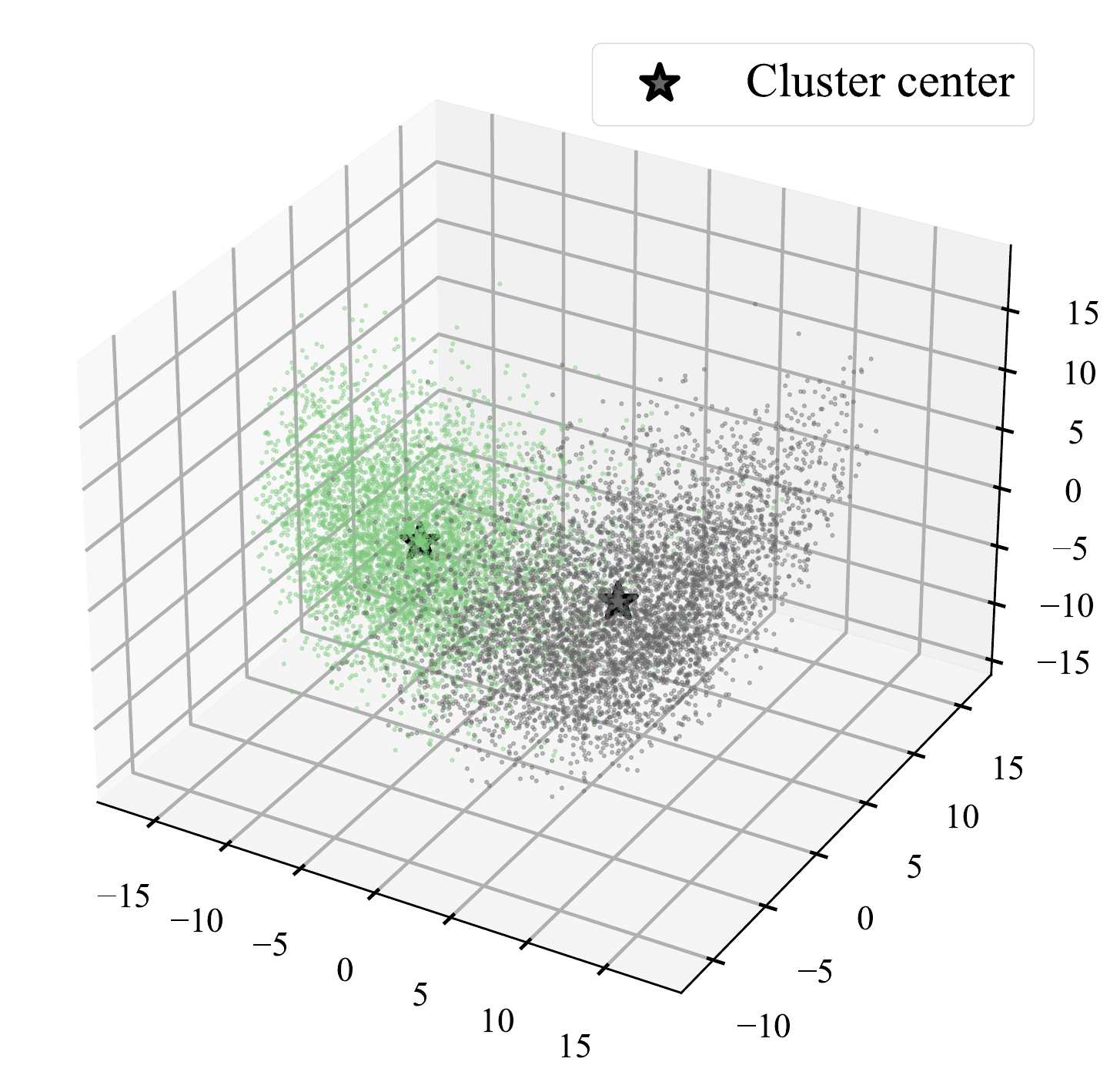}} 
\caption{3D PCA visualization of Amazon dataset in different sentence embedding spaces. (a) Two classes are clearly separated, and we can even find sub-topics by clustering texts in the same class. We can even observe a clear linear semantic structure. (b) Data sub-structures are somewhat kept, but the two sentiment classes are not separated distinctly. (c) The two classes are better distinguished, but the detailed data structures are lost. }
\label{fig:vis}
\end{figure*}

\paragraph{When does \texttt{SimPTC} not work very well?} Both \texttt{Vanilla Prompting} and \texttt{E\&M} suffer when the class names are abstract concepts, e.g., subjective and objective in the Subj dataset. This suggests that prompting and current text embeddings are still poor at linking texts to class names describing abstract properties. But interestingly, the two classes of Subj separate well in the SimCSE embedding space (Figure \ref{fig:tsne_subj}), indicating the ability of PLM embedding spaces to capture abstract semantic concepts. Additionally, both methods underperform self-training keyword-based methods in long document tasks (see Appendix \ref{sec:ap_tc14_analysis} for more details).

\subsection{Different Encoders}\label{sec:encoder}

In this section, we utilize \texttt{SimPTC} to analyze different PLM embedding spaces. Specifically, we ask two questions: 1) \textbf{Are the texts also clustered by topics in the embedding spaces of PLMs that are not explicitly trained to generate meaningful embeddings?} 2) \textbf{Are the embeddings of larger PLMs more informative?} To answer these questions, we compare RoBERTa\sslarge~(\texttt{RL}) \citep{liu2019roberta}, Sentence RoBERTa\sslarge~(\texttt{SRL}) \citep{reimers-gurevych-2019-sentence}, SimCSE supervised RoBERTa\sslarge~(\texttt{SimCSE}), and \texttt{T5-3B} \citep{raffel2020exploring} embedding spaces. For sentence embeddings, we average the embeddings of all tokens in a text for \texttt{RL}, and use the embeddings of the last hidden states from the encoder for \texttt{T5-3B}.
\subsubsection{Quantitative Results} \label{sec:diff_enc_quant}

\paragraph{PLM embeddings can categorize text without task-specific fine-tuning.} As \texttt{RL} is not pre-trained to generate meaningful sentence embeddings, \texttt{E\&M} does not work with \texttt{RL}. So we initialize \texttt{SimPTC} using \texttt{Vanilla Prompting}. We do the same for \texttt{SRL} as it provides a better initialization. We share the covariance matrices to offer extra regularization. Surprisingly, as shown in Table \ref{tab:encoder}, the original \texttt{RL} achieves comparable performance to \texttt{SimCSE} and outperforms the more sophisticated sentence encoder \texttt{SRL} on 4 out of 5 datasets.

\paragraph{Larger PLMs tend to have more informative embedding spaces.} With a larger model \texttt{T5-3B}, \texttt{SimPTC} gives even better results. Initialized using \texttt{VP}, \texttt{T5-3B} achieves comparable or better performance on 4 out of 5 datasets than the state-of-the-art sentence encoder \texttt{SimCSE}, matching even the supervised \texttt{BERT} performance on IMDb. This indicates that embedding spaces of larger PLMs might have even better clustering properties, which agrees with their stronger zero-/few-shot learning ability.

\begin{table}
\centering
\small
\begin{tabular}{lllllll}
\toprule
\textbf{Encoder}& \textbf{Size} & \textbf{AG} & \textbf{DB} & \textbf{YH} & \textbf{AM} & \textbf{IM}\\
\midrule
\texttt{SimCSE}         & 350M  & $\textbf{86.9}$ & $93.2$ & $\textbf{63.9}$ & $93.9$ & $91.0$\\
\texttt{RL}$\dagger$    & 350M  & $86.1$ & $96.0$ & $54.2$ & $93.5$ & $92.3$\\
\texttt{SRL}$\dagger$   & 350M  & $85.8$ & $93.4$ & $55.4$ & $92.9$ & $90.9$\\
\texttt{T5-3B}$\dagger$ & 3B  & $86.7$ & $\textbf{96.7}$ & $55.1$ & $\textbf{95.3}$ & $\textbf{94.5}$\\
\midrule
\texttt{BERT}(sup.)               & 110M  & $94.4$  & $99.4$ & $75.0$ & $97.2$ & $94.5$\\
\bottomrule
\end{tabular}
\caption{\label{tab:encoder}
Comparison of different encoders. $\dagger$: Clusters are initialized using \texttt{Vanilla Prompting} (\S\ref{sec:diff_enc_quant}).
}
\end{table}
\subsubsection{Qualitative Analysis} 
To explain the first finding in \S\ref{sec:diff_enc_quant}, we analyze the 3D PCA visualization of the Amazon dataset in three embedding spaces (Figure \ref{fig:vis}). We observe that: 1) \texttt{RL} preserves the dataset sub-structures, but the two sentiment clusters do not separate very well. 2) \texttt{SRL} pushes semantically close texts together by introducing an extra training objective, which leads to more separable clusters, but the detailed structures of data are lost. 3) The SimCSE embeddings separate the two classes distinctively, and the texts are further clustered together by sub-topics, such as books or products. Very interestingly, a clear linear semantic sub-structure can be observed:
\begin{align*}
    \bar{\mb v}_{pos}^{book}-\bar{\mb v}_{neg}^{book}&\approx \bar{\mb v}_{pos}^{prod}-\bar{\mb v}_{neg}^{prod} \approx \bar{\mb v}_{pos}-\bar{\mb v}_{neg},
\end{align*}
where $\bar{\mb v}_{neg}^{prod}$ is the cluster center vector of all negative product reviews; $\bar{\mb v}_{pos}$ and $\bar{\mb v}_{neg}$ are the centers of two sentiment classes. Therefore \texttt{RL} outperforms \texttt{SRL} possibly because it is more descriptive of texts. With a good separability of topics and the ability to capture data sub-structures, \texttt{SimCSE} achieves the best overall zero-shot classification performance.
\section{Conclusion}\label{conclustion}

In this work, we show that a simple clustering-based approach, \texttt{SimPTC}, can achieve state-of-the-art zero-shot text classification performance on a wide range of tasks. With extensive experiments, we identify the keys to cluster texts in the PLM embedding spaces and also the limitations of \texttt{SimPTC}. Further analysis of different PLMs shows that PLMs can categorize texts in their embedding spaces without being trained to derive semantically meaningful sentence embeddings, and Larger PLMs tend to have more informative embeddings. We hope our exploration into the embedding spaces of PLMs can provide insights into understanding and developing new methods to elicit the zero-/few-shot learning ability of PLMs. 

\section*{Limitations}

We identify three limitations of \texttt{SimPTC} as well as this work: 1) Due to the nature of clustering and sentence embeddings, \texttt{SimPTC} still suffers at many-class tasks with long documents and tasks with abstract class names (e.g., subjective v.s. objective); 2) Currently applying \texttt{SimPTC} to other NLP tasks like NLI is not straightforward. 3) Due to computational resource constraints, our analysis is limited to PLMs with parameters up to 3 Billion. It would be interesting to see if our observations generalize to the largest models like GPT-3 (175B) \citep{NEURIPS2020_1457c0d6} and PaLM (540B) \citep{chowdhery2022palm}, which show the strongest zero-/few-shot ability.

\section*{Ethics Statement}


This work aims to analyze how to use PLM knowledge in their embedding spaces to categorize texts on different topics. 
Unlike many other deep-learning-based models, \texttt{SimPTC} involves no large neural model pre-training, re-training, or fine-tuning throughout the entire development of the method. Once we get the embeddings of the unlabeled texts, the PLMs are not used anymore. Thus developing and applying our approach requires only minimal computational resources and cause fewer carbon emissions than methods that require dataset-specific fine-tuning or engineering. Besides, we do not anticipate any signiﬁcant ethical issues introduced by our approach. We use only off-the-shelf PLMs, and the datasets involved are all publicly available topic or sentiment classification datasets. Nevertheless, we urge anyone to evaluate the robustness of the method before using \texttt{SimPTC} in sensitive contexts such as healthcare or legal scenarios.

\bibliography{anthology,custom}
\bibliographystyle{acl_natbib}

\clearpage
\appendix

\section{Expanded Class Name Examples for All Datasets}\label{sec:ap_class_names}
Some examples of the original and extracted class names are shown in Table \ref{tab:ap_class_names_ag} - \ref{tab:ap_class_names_sent}.

\section{Templates Used for All Datasets}\label{sec:ap_templates}
\quad AG's News:
\begin{tcolorbox}
$\begin{aligned}
    &\text{The news is about }\langle mask \rangle.\\
    &\text{The news is related to }\langle mask \rangle.\\
    &\langle mask \rangle\text{ is the topic of the news}.\\
    &\text{This week's news is about }\langle mask \rangle.
\end{aligned}
$
\end{tcolorbox}
DBpedia:
\begin{tcolorbox}
$\begin{aligned}
    &\text{The object is about }\langle mask \rangle.\\
    &\text{The object is related to }\langle mask \rangle.\\
    &\langle mask \rangle\text{ is the topic of the object}.\\
    &\langle mask \rangle\text{ is the subject of the object}.
\end{aligned}
$
\end{tcolorbox}
Yahoo:
\begin{tcolorbox}
$\begin{aligned}
    &\text{The answer is about }\langle mask \rangle.\\
    &\text{The answer is related to }\langle mask \rangle.\\
    &\langle mask \rangle\text{ is the topic of the answer}.\\
    &\langle mask \rangle\text{ is involved in the answer}.
\end{aligned}
$
\end{tcolorbox}
Amazon:
\begin{tcolorbox}
$\begin{aligned}
    &\text{A }\langle mask \rangle \text{ product review}.\\
    &\text{The product review is }\langle mask \rangle.\\
    &\text{The reviewer found the product }\langle mask \rangle.\\
    &\text{The product is }\langle mask \rangle.\\
\end{aligned}
$
\end{tcolorbox}
IMDb:
\begin{tcolorbox}
$\begin{aligned}
    &\text{A }\langle mask \rangle \text{ movie review}.\\
    &\text{The movie review is }\langle mask \rangle.\\
    &\text{The reviewer found the movie }\langle mask \rangle.\\
    &\text{The movie is }\langle mask \rangle.\\
\end{aligned}
$
\end{tcolorbox}
\begin{table*}
\centering
\begin{tabular}{ccccccc}
\toprule
\textbf{Name} & \textbf{Type} & \textbf{\# Class} & \textbf{Training Size} & \textbf{Test Size}& \textbf{Max Iter} & \textbf{Covariance Setting} \\ 
\midrule
AG's News   & Topic     & 4     & 120000    & 7600  & 50 & Full\\
DBpedia     & Topic     & 14    & 560000    & 70000 & 40 & Full\\
Yahoo       & Topic     & 10    & 1400000   & 60000 & 20 & Full\\
Amazon      & Sentiment & 2     & 200000    & 10000 & 50 & Tied\\
IMDb        & Sentiment & 2     & 25000     & 25000 & 150 & Tied\\
\bottomrule
\end{tabular}
\caption{Statistics of datasets used to compare with state-of-the-art methods in \S\ref{sec:sota} and extra model settings. \texttt{SimPTC} stops when the model prediction stops changing, or the maximum number of iteration is achieved. Full: each Gaussian mixture has its own covariance $\bSigma_k$. Tied: all Gaussians share the same covariance $\bSigma$.}
\label{tab:ap_dataset_model}
\end{table*}
\section{Math foundation of the \texttt{SimPTC}Update Step}\label{sec:ap_update}
Bayesian approaches inject prior knowledge by introducing prior distribution on model parameters while still allowing the model to fit the data. In this section we first discuss the prior distributions we choose. Then we show how these choices affect the model prediction by analyzing the maximum a posteriori probability (MAP) solution of model parameters.
\paragraph{Prior distributions} Following \citet{bishop2006pattern}, we choose a Dirichlet distribution as the prior for mixture weights $\bpi$, and a Gaussian-Wishart prior for the mean and precisions, i.e., the inverse of covariance $\bLambda=\bSigma^{-1}$:
\begin{align*}
    p(\bpi) =&~\text{Dir}(\bpi|\alpha_0)=C(\alpha_0)\prod_k \pi_k^{\alpha_0-1}\\
    p(\bmu,\bLambda)=&~p(\bmu|\bLambda)p(\bLambda)\\
    =&~\prod_k \calN(\bmu_k|\mb m_0, (\beta_0\bLambda_k)^{-1})\cdot \\
    &~\mathcal{W}(\bLambda_k|\mb W_0,\nu_0),
\end{align*}
where $C(\alpha_0)$ is a normalizing constant, and $\alpha_0$ can be interpreted as the prior number of observations associated with each mixture. We simply choose $\alpha_0=\frac{N}{K}$ to favor balanced weights. For the means and covariances, we offer the model maximum freedom to fit the data by choosing a non-informative prior \citep{murphy2007conjugate}. Specifically, we set:
\begin{equation}\label{eq:prior}
    \mb m_0=0,\ \beta_0\rightarrow0,\ \mb W_0 = \frac{1}{d}\bSigma_{init}^{-1},\ \nu_0 = d,
\end{equation}
where $\bSigma_{init}$ is some initial guess of the covariance matrix, which can be set as the empirical covariance of the data. Then we update the model with the standard variational optimization \citep{bishop2006pattern} for Bayesian GMM. 
\paragraph{MAP solution} Here, we show the MAP solution after one update step to give some intuition about how our choice of prior model parameters \eqref{eq:prior} influences the update of model parameters. As the standard EM update of maximum likelihood methods, the variational update also contains two steps. In the variational E step, we evaluate the responsibilities using the current variational distribution parameters:
\begin{equation*}
    r_{nk}:=\mathbb{E}_{\bpi,\bmu,\bSigma}[z_{nk}],
\end{equation*}
where $z_{nk}$ is the binary latent variable indicating whether data $x_n$ belongs to cluster $k$; and in the variational M step, we update the variational distribution parameters. For simplicity, we introduce the following statistics:
\begin{align*}
    N_k& =\sum_n r_{nk}\\
    \bar{\mb x}_k&=\frac{1}{N_k}\sum_n r_{nk}\mb x_n\\
    \mb S_k &= \frac{1}{N_k}\sum_n r_{nk}(\mb x_n-\bar{\mb x}_k)(\mb x_n-\bar{\mb x}_k)^\top.
\end{align*}
Then the MAP solution of $\bpi,\bmu,\bSigma$ given the responsibilities $r_{nk}$'s after a variational M steps is given by
\begin{equation}\label{eq:map}
    \begin{aligned}
            \pi_k^* &=\frac{\alpha_0 -1 + N_k}{K(\alpha_0-1) + N}\\
    \bmu_k^* &= \bar{\mb x}_k\\
    \bSigma^*_k &=\frac{d\bSigma_{init} + N_k \mb S_k}{N_k-1},
    \end{aligned}
\end{equation}
where $d$ is the number of feature dimensions and K is the number of classes. We can see that by choosing non-informative prior \eqref{eq:prior} of $(\bmu, \bSigma)$, we allow the model to fit the data with maximum freedom. By choosing a large $\alpha_0$, we can push the mixing weights towards uniform but still allow the model to fit the data.

\section{Datasets Statistics and Model Settings}\label{sec:ap_dataset}
The statistics of the five datasets used in \S\ref{sec:sota} and max iteration numbers can be found in Table \ref{tab:ap_dataset_model}. For Amazon, we use the same test set sampled by \citet{hu2021knowledgeable} and randomly sample 200,000 texts from the original training set for the unlabeled training data. Since SimCSE only handles texts with a maximum length of 512, we crop texts with lengths exceeding 512. We choose the maximum number of iterations empirically according to the size of the unlabeled data which is equal to the training set size plus the test set size. For topic datasets, each Gaussian has its individual covariance matrix. For sentiment datasets, all Gaussians share the same covariance matrix to provide extra regularization as the data is relatively sparse. The effect of sharing the covariance matrix is discussed in Appendix \ref{sec:ap_covariance}.
\begin{figure}[t]
\centering
\includegraphics[width=\columnwidth]{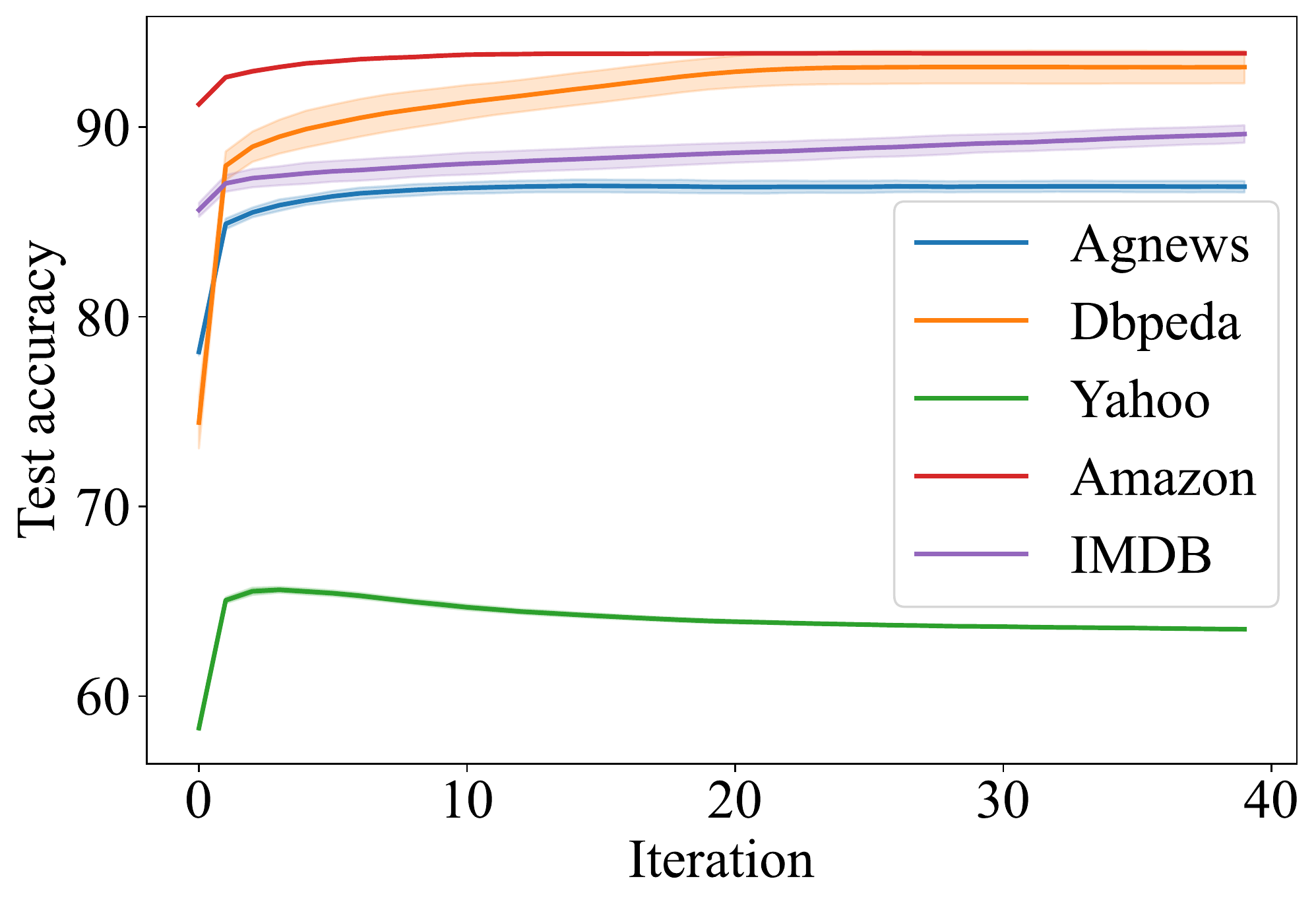}
\caption{The performance v.s. update iteration plot of \texttt{SimPTC} on all five datasets. The solid line shows the average accuracy at each iteration, whereas the blurred area indicates the standard deviation of using different templates. \textbf{\texttt{SimPTC} converges to a good-quality prediction as the clustering process converges.}}
\label{fig:converge}
\end{figure}
\section{Convergence Analysis}\label{sec:ap_converge}
Although \texttt{SimPTC} is guaranteed to converge, it is unclear whether it will converge to a good solution when the algorithm stops. Therefore we study how the model performance varies as the updating process proceeds. We plot the test accuracy of intermediate update steps on all datasets in Figure \ref{fig:converge}, where the standard deviations caused by using different templates are illustrated with blurred areas. We observe that the performance gradually improves and converge in all dataset except Yahoo, where \texttt{SimPTC} still converges to a result much better than the initialization. Also, as shown in the blurred areas in Figure \ref{fig:converge}, the update step is stable when different templates are used. Moreover, \texttt{SimPTC} almost converges on all five datasets under our setting of the maximum number of iterations.

\begin{figure}[t]
\centering
\includegraphics[width=\columnwidth]{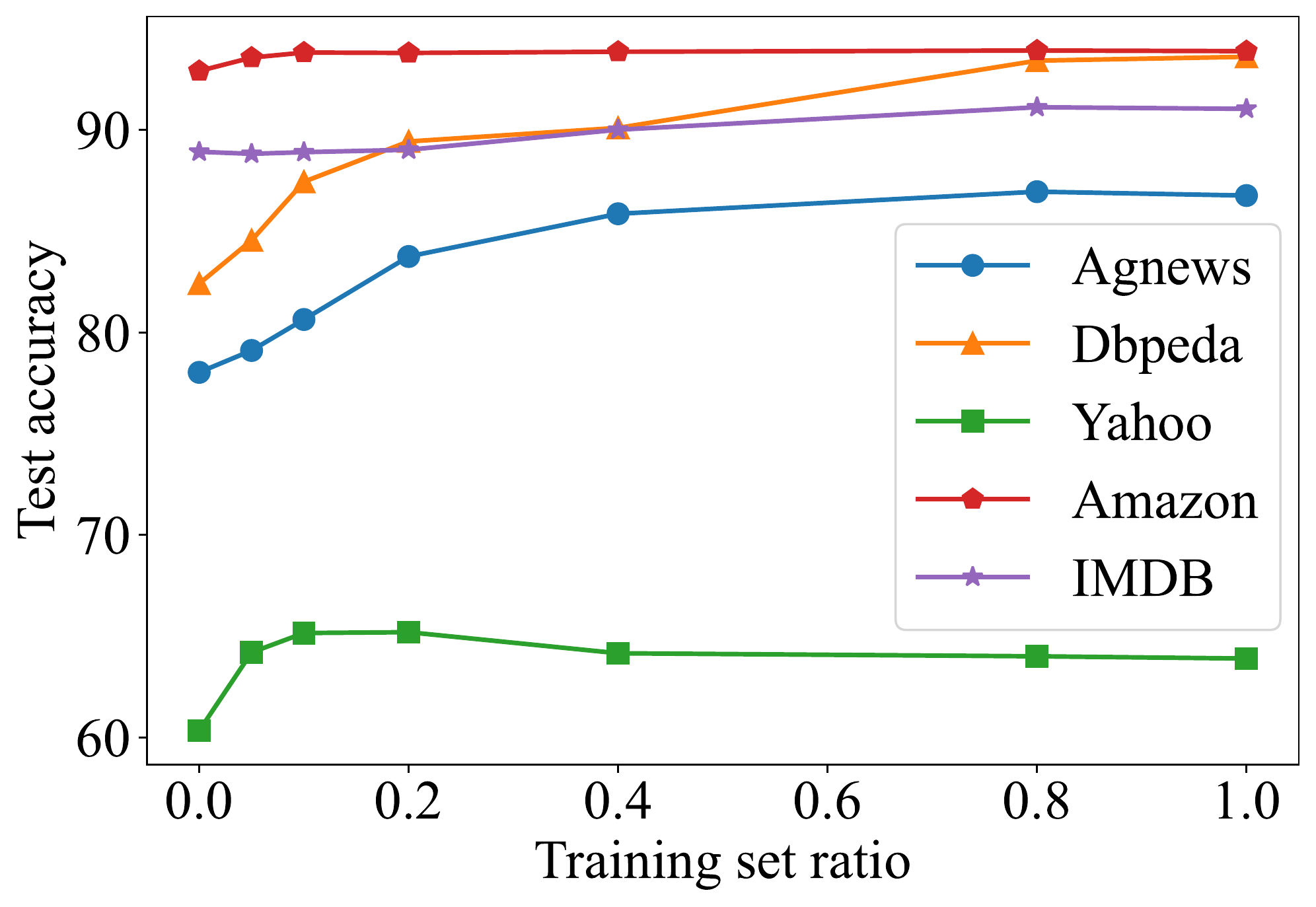}
\caption{The performance v.s. unlabeled dataset size plot of \texttt{SimPTC} on all five datasets. More unlabeled data, in general, tends to improve the prediction of \texttt{SimPTC}.}
\label{fig:dataset}
\end{figure}
\section{Effect of Unlabeled Dataset Size}\label{sec:ap_dataset_size}
In the standard setting, we use both the train and test set for fitting the Bayesian GMM. To study the effect of the unbalanced dataset size, we keep the unlabeled test data and use the training data with ratios varying from $0$ to $1$. As illustrated in Figure \ref{fig:dataset}, on almost all datasets, more unlabeled data brings more improvement. 

One possible explanation is: to model the shape of all clusters with a certain error threshold, one needs samples of  a number at least linear to the number of dimensions \citep{vershynin2012close} and linear to the number of classes. Therefore a large unlabeled dataset helps the model to fit data with many classes in a high-dimensional space better (for RoBERTa\sslarge~, the number is $1024$). By sharing the covariance matrix (Amazon and IMDb), we reduce the number of model parameters. Thus \texttt{SimPTC} works better than fitting individual covariance for each cluster (Agnews, Dbpedia, and Yahoo) when the data is sparse. Since for many tasks collecting unlabeled data is considered to be much easier than collecting annotated data, we can improve the performance of \texttt{SimPTC} in real-world applications  at a low cost. 

\section{Effect of Sharing Covariance Matrix}\label{sec:ap_covariance}
We explore two covariance settings in \texttt{SimPTC}. \emph{Full}: each Gaussian mixture has its own covariance $\bSigma_k$, and \emph{tied}: all Gaussians share the same covariance $\bSigma$. Note that the sharing the covariance matrices (the \emph{full} and \emph{tied} setting) is a standard hyperparameter of GMM. The \emph{full} setting is more flexible, and as Table \ref{tab:cov} shows it improves the initial \texttt{E\&M} predictions on all datasets. By sharing the covariance matrices (the \emph{tied} setting) we 1) reduce model parameters to provide extra regularization and 2) add stronger assumptions on the cluster shapes. Therefore it is useful when
\begin{itemize}
    \item the data is relatively sparse (e.g., IMDb in Table \ref{tab:cov} and TC14 datasets in \S\ref{sec:tc14}),
    \item the embedding space of PLM is less structured (T5 and RoBERTa\sslarge~embeddings (\S\ref{sec:encoder})),
    \item the texts of different classes describe similar objects (e.g., sentiment tasks).
\end{itemize}
Otherwise, we recommend allowing clusters to have different covariances.


\begin{table}
\centering
\small
\begin{tabular}{lccccc}
\hline
\multirow{2}{*}{\textbf{Setting}} & \multicolumn{3}{c}{\bf Topic}& \multicolumn{2}{c}{\bf Sentiment} \\ 
\cline{2-6}
 & \textbf{AG}& \textbf{DB}& \textbf{YH}& \textbf{AM}& \textbf{IM} \\ 
	\hline
\texttt{E\&M} & $78.2 $ & $ 74.4 $ & $58.3 $ & $91.2 $ & $85.6 $\\
Full & $\bf 86.9\textcolor{green}{ \uparrow}$ &   $\bf 93.2\textcolor{green}{ \uparrow}$&	$\bf 63.9\textcolor{green}{ \uparrow}$&	$92.4\textcolor{green}{ \uparrow}$&	$86.2\textcolor{green}{ \uparrow}$\\
Tied & $86.5\textcolor{green}{ \uparrow}$ &	$90.8\textcolor{green}{ \uparrow}$&	$56.9\textcolor{red}{\downarrow}$&	$\bf 93.9\textcolor{green}{  \uparrow}$&	$\bf 91.0\textcolor{green}{ \uparrow}$\\
\hline
\end{tabular}
\caption{Average test accuracy of all templates with different covariance settings. Tied: all Gaussians share the same covariance matrix. Full: every Gaussian has its own covariance matrix.
}
\label{tab:cov}
\end{table}
\begin{table*}
\centering
\begin{tabular}{cccc}
\toprule
\textbf{Name} & \textbf{Type} & \textbf{Class name examples} & \textbf{Template for prompting} \\ 
\midrule
20 News	    &Topic  &comp.graphics; sci.space &[ Category : $\langle mask\rangle$] $\langle text\rangle$\\
NYT-Topic	&Topic	&business; politics; sports 	&[ Category : $\langle mask\rangle$] $\langle text\rangle$\\
NYT-Location&location	&united\_states; iraq; japan	&[ Category : $\langle mask\rangle$] $\langle text\rangle$\\
BBC News	&Topic	&sport; business; entertainment	&[ Category : $\langle mask\rangle$] $\langle text\rangle$\\
Emotion&	Emotion	&sad; joy; anger	&[ Category : $\langle mask\rangle$] $\langle text\rangle$\\
Banking77	&Intent	&activate\_my\_card; age\_limit &[ Category : $\langle mask\rangle$] $\langle text\rangle$\\
TREC	&Question	&abbr.; entity; description	&[ Category : $\langle mask\rangle$] $\langle text\rangle$\\
Biomedical	&Paper title	&aging; chemistry; erythrocytes	&[ Category : $\langle mask\rangle$] $\langle text\rangle$\\
StackOverﬂow	&Question	&svn; oracle; bash	&[ Category : $\langle mask\rangle$] $\langle text\rangle$\\
Yelp	    &Sentiment	&positive; negative	&It is $\langle mask\rangle$. $\langle text\rangle$\\
SST-2	&Sentiment	&positive; negative	&It is $\langle mask\rangle$. $\langle text\rangle$\\
SST-5	&Sentiment &very positive; positive; negative	&It is $\langle mask\rangle$. $\langle text\rangle$\\
MPQA	&Opinion polarity	&positive; negative	&It is $\langle mask\rangle$. $\langle text\rangle$\\
Subj	&Subjectivity	&subjective; objective	&It is $\langle mask\rangle$. $\langle text\rangle$\\
\bottomrule
\end{tabular}
\caption{Extra information about TC14 datasets. Template for prompting is the template we used to perform prompt-based zero-shot learning, i.e., \texttt{Vanilla Prompting}. We use the same template for all sentiment tasks and another for all other datasets.}
\label{tab:ap_tc14}
\end{table*}
\section{Implicit Balanced Assumption of KPT}\label{sec:ap_kpt}
\citet{hu2021knowledgeable} proposed a data-dependent \texttt{Contextualized Calibration (CC)}. They motivate \texttt{CC} by observing that some label words are less likely to be predicted than others, regardless of the label of input sentences. To solve the problem, \texttt{CC} works in the following steps: First, to estimate a contextualized prior distribution of label words using some sampled unlabeled data:
    \begin{equation}\label{eq:cp}
        \begin{aligned}
            P_{\mathcal{D}}(v)&=\mathbb{E}_{{\bf x}\sim \mathcal{D}}P_{\mathcal{M}}([MASK]=v|{\bf x})\\
            &\approx\frac{1}{|C|}\sum_{{\bf x}\in C}P_{\mathcal{M}}([MASK]=v|{\bf x}),
        \end{aligned}
    \end{equation}
where $v$ stands for a particular label word, $\mathcal{D}$ is the data distribution, $P_{\mathcal{M}}$ is the model prediction, $C$ is a sampled subset of the dataset. Then they use the contextualized prior of label words to calibrate the predicted distribution:
    \begin{equation}\label{eq:contextual}
        \tilde{P}_{\mathcal{M}}([MASK]=v|{\bf x}) \propto \frac{P_{\mathcal{M}}([MASK]=v|{\bf x})}{P_{\mathcal{D}}(v)}.
    \end{equation}
The final probability is normalized to 1. 

The contextualized prior can be interpreted as a marginal distribution. Consider we have one label word for each class. The contextualized prior measures the portion of each class in the dataset based on the model's predictions. Then \texttt{CC} penalizes the probability of predicting one class if the model thinks it assigns too many samples to this class ($P_{\mathcal{D}}(v)$ is large). Intuitively this is to force the model to assign equal numbers of samples to each class, which is to force a uniform marginal distribution. The underlying implicit assumption is that the dataset is balanced. Although \texttt{CC} improves the zero-shot performance of \texttt{KPT}, we argue that this is because the evaluation datasets happen to be balanced, and \texttt{CC} becomes problematic when the dataset is unbalanced (see C2 in \S\ref{sec:sota}).

\begin{table*}
\centering
\small
\begin{tabular}{llllllll}
\toprule
 \textbf{Method} & \textbf{20News} &\textbf{NYT-T} &\textbf{NYT-L} &\textbf{BBC} &\textbf{Yelp} &\textbf{Emotion} &\textbf{Banking77} \\
 \midrule
 \texttt{VP} &41.0/36.6&	\textbf{72.1}/55.5&	66.3/62.0&	75.8/73.8&	80.6/80.0&	21.7/19.3&	21.2/16.5\\
 \texttt{Encode\&Match}     &42.9/42.0&	59.6/53.9&	65.9/66.4&	80.6/80.4&	93.3/93.3&	\textbf{52.3}/46.2&	57.0/55.6\\
 \texttt{SimPTC}            &\textbf{51.2}/\textbf{53.1}&	66.0/\textbf{63.5}&	\textbf{72.1}/\textbf{77.7}&	\textbf{89.5}/\textbf{89.7}&	\textbf{94.3}/\textbf{94.3}&  51.0/\textbf{46.6}&	\textbf{66.6}/\textbf{66.7}\\
 \midrule
 Zero-shot SOTA & 78.6/77.8$^a$ & 79.0/68.6$^a$ & 91.8/92.0$^a$ & 84.0 (acc)$^b$ & 90.0/90.0$^a$ & -/- & -/33.2$^c$\\
 \midrule
 &\textbf{SST-2} &\textbf{SST-5} &\textbf{MPQA} &\textbf{Subj} &\textbf{TREC} &\textbf{Biomed.} &\textbf{StackOF}\\
  \midrule
  \texttt{VP}&73.7/72.0&	32.4/28.4&	49.0/48.6&	\textbf{56.2}/47.8&	\textbf{37.7}/28.1&	25.0/22.4&	26.6/21.0\\
  \texttt{Encode\&Match}    &82.0/81.9&	42.7/38.7&	83.8/82.4&	51.7/47.2&	35.4/28.9&	26.8/24.7&	49.0/49.4\\
  \texttt{SimPTC}           &\textbf{86.8}/\textbf{86.8}&	\textbf{46.2}/\textbf{42.2}&	\textbf{84.8}/\textbf{83.4}&	53.9/\textbf{52.0}&	37.3/\textbf{30.8}&	   \textbf{38.4}/\textbf{40.6}&	\textbf{74.2}/\textbf{77.9}\\
  \midrule
 Zero-shot SOTA  &83.6 (acc)$^d$   &35.0 (acc)$^d$    &67.6 (acc)$^d$ &51.4 (acc)$^d$ & 32.0 (acc)$^d$ & 46.2 (acc)$^e$ & 75.5 (acc)$^e$ \\
\bottomrule
\end{tabular}
\caption{\label{tab:other_datasets}
Zero-shot micro-/macro-F1 scores on other datasets. \texttt{VP}: vanilla prompting (\S\ref{sec:experiment}). We collect publicly available zero-shot state-of-the-art (SOTA) method performance as a reference. a: X-Class, \citep{wang-etal-2021-x} a SOTA keyword-based method. b: \citep{harrando2021explainable}. c: Crowdsourced human performance from \citet{alex2021raft} (they used a selected portion of Banking77). d: zero-shot prompt-based zero-shot learning provided by \citet{gao-etal-2021-making}. e: SCCL, a contrastive-learning-based unsupervised text clustering method by \citet{zhang2021supporting}. SCCL forces on clustering texts of different topics. When calculating accuracy, the labels of clusters are determined by solving a min-cost perfect matching problem based on the predicting accuracy.
}
\end{table*}
\section{TC14 Datasets}\label{sec:ap_tc14}
To study the applications and limitations of \texttt{SimPTC}, we collect the following 14 datasets with diverse topics, text lengths, and class numbers. Specifically, we did a literature search in zero-shot text classification and collected datasets that best fit our text classification setting with label names that have class-info. We first introduce the details of the TC14 datasets (\S\ref{sec:ap_tc14_info}). Then we discuss the implementation details in \S\ref{sec:ap_tc14_implementation}. We show the full results in \S\ref{sec:ap_tc14_results} and provide extra analysis in \S\ref{sec:ap_tc14_analysis}.
\subsection{Dataset Information}\label{sec:ap_tc14_info}
The datasets we used are:
\begin{itemize}[topsep=0pt,itemsep=0pt,partopsep=0pt, parsep=0pt, leftmargin=*] 
    \item \textbf{20 News} \citep{lang1995newsweeder} is a news classification dataset. It has a relatively long average text length and many classes.
    \item \textbf{NYT-Topic} \citep{10.1145/3366423.3380278} is a long document topic classification dataset  that is very unbalanced.
    \item \textbf{NYT-Location} \citep{10.1145/3366423.3380278} uses the same corpus as NYT-Topic but categorizes the texts according to locations. The dataset is very unbalanced.
    \item \textbf{BBC News} \citep{10.1145/1143844.1143892} is a news dataset containing 2225 articles.
    \item \textbf{Yelp} \citep{NIPS2015_250cf8b5} is a review sentiment dataset.
    \item \textbf{Emotion} \citep{saravia-etal-2018-carer} is a dataset of English Twitter messages with six basic emotions, and the dataset is very unbalanced.
    \item \textbf{Banking77} \citep{Casanueva2020} is a dataset composed of online banking queries annotated with their corresponding intents. It has a very fine-grained set of intents in the banking domain. 13,083 customer service queries are categorized into 77 intents.
    \item \textbf{SST-2} \citep{socher-etal-2013-recursive} is a sentence sentiment classification dataset.
    \item \textbf{SST-5} \citep{socher-etal-2013-recursive} is a fine-grained sentiment classification dataset. Texts are classified into five sentiment classes: very negative, negative, neutral, positive, and very positive.
    \item \textbf{MPQA} \citep{wiebe2005annotating} is an opinion polarity analysis dataset.
    \item \textbf{Subj} \citep{pang2004sentimental} is a subjectivity analysis dataset. 
    \item \textbf{TREC} \citep{voorhees2000building} is an unbalanced question classification dataset.
    \item \textbf{Biomedical} \citep{xu2017self} is a paper title classification dataset, where 20,000 titles are categorized into 20 groups.
    \item \textbf{StackOverﬂow} \citep{xu2017self} is a dataset containing 20,000 questions with 20 classes.
\end{itemize}
Since we are evaluating zero-shot methods, we report scores on the full datasets (dataset sizes are shown in Table \ref{tab:other_datasets_stat}).
\subsection{Additional Implementation Details}\label{sec:ap_tc14_implementation}
We compare with \texttt{Vanilla Prompting} rather than \texttt{KPT} because KPT has an improper balanced dataset assumption (\S\ref{sec:sota} C3), and \texttt{KPT} cannot handle class names containing multiple words. 
\par For the 20 News dataset, we use class names from \citet{mekala-shang-2020-contextualized} as the original class names are not complete English. We implement \texttt{Vanilla Prompting} using OpenPrompt \citep{ding2021openprompt}. When a class name contains multiple words, we use the average probability of predicting each word as implemented in OpenPrompt. BBC News contains only 2225 texts and is too small to fit a 1024-by-1024 covariance matrix even if we share the covariance matrices of clusters. Banking77 has too many classes compared with the dataset size, and as a result, \texttt{Encode\&Match} assing zero samples to some classes. To fix these two problems, we perform a PCA to reduce the feature dimension such that the reconstruction error is 3\% before \texttt{Encode\&Match}.
\subsection{Full Results}\label{sec:ap_tc14_results}
We report the micro-macro F1 scores on TC14 in Table \ref{tab:other_datasets}. For comparison, we also collect publicly available state-of-the-art results on these datasets. Some papers only report the accuracy of their models, and we report these numbers instead.
\subsection{Additional Analysis}\label{sec:ap_tc14_analysis}

As discussed in \S\ref{sec:tc14}, both prompting and \texttt{E\&M} suffer on the Subj dataset where the class names are abstract concepts (subjective v.s. objective). As a result, \texttt{SimPTC} also does not go very far from random guessing (50\%). However, despite \texttt{E\&M} failing to link the texts correctly with the abstract class names, the texts themselves are well-separated in the embedding space (Figure \ref{fig:tsne_subj}). This suggests that texts with abstract classes can also be clustered together in the PLM embedding spaces. A 10-shot setting (averaged over 5 seeds) improves \texttt{SimPTC} from 52.0 to 89.2 on Subj, outperforming GPT-3 175B in-context learning (76.4).

In terms of limitations, another important observation is that: on long document classification tasks (20 News, NYT-Topic, NYT-Location), both \texttt{SimPTC} and \texttt{Vanilla Prompting} underperform the state-of-the-art keyword-based method X-Class \citep{wang-etal-2021-x}, showing an information loss when PLMs encodes long documents into the embedding spaces. This indicates that in terms of extracting information from long documents, self-training keyword-based approaches still perform better than zero-shot our clustering-based approach and prompting methods.

\begin{table*}
\centering
\begin{tabular}{cc}
\toprule
\textbf{Class Name} & \textbf{Expanded Class Names}  \\ 
\midrule
\multirow{2}{*}{politics}   & alt rightist, social fascism, psychopolitical, leader of opposition, junior minister,\\
& whipped vote, political, regressive leftism, policy making, dollar democracy, ...\\
\midrule
\multirow{2}{*}{sports}   & professional baseball, game set match, banana ball, empty bench, first touch,\\
& football, sportsman, visiting team, athletic, exhibition game, super cup, ...\\
\midrule
\multirow{2}{*}{business}   & account name, commerciality, making money, sprinkler strategy, web company,\\
& consumer good, business economics, maintained markup, commercial enterprise, ...\\
\midrule
\multirow{2}{*}{technology}   & cryoengineering, aeronautical engineering, geotechnology, cwm silicon, nuclearism,\\
& digital technology, cryotechnology, xenotechnology, applied science, deepfake, ...\\
\bottomrule
\end{tabular}
\caption{Original class names and expanded class names of AG's News.}
\label{tab:ap_class_names_ag}
\end{table*}

\begin{table*}
\centering
\begin{tabular}{cc}
\toprule
\textbf{Class Name} & \textbf{Expanded Class Names}  \\ 
\midrule
\multirow{2}{*}{company}   & hook stock, private corporation, large company, big company, business organization,\\
& furniture company, companies, sprinkler strategy, corp, livery company, ...\\
\midrule
\multirow{2}{*}{school}   & elementary schooler, undergraduates, university student, dual school, antiuniversity,\\
& schoolless, overschooled, secondary modern, science room, state school, ...\\
\midrule
\multirow{2}{*}{artist}   & arte povera, ernstian, art show, da vincian, polystylist, gallery opening, pricasso,\\
& artworks, artistdom, superrealist, artists, clean brushes, post impressionist ...\\
\midrule
\multirow{2}{*}{athlete}   & olga korbut, athleticism, pull muscle, walking sports event, pancratical,nongymnast,\\
& sportswomen, athletic contest, weightlifter, winter olympics competition, ...\\
\midrule
\multirow{2}{*}{politics}   & alt rightist, social fascism, psychopolitical, leader of opposition, junior minister,\\
& whipped vote, political, regressive leftism, policy making, dollar democracy, ...\\
\midrule
\multirow{2}{*}{transportation}   & antirail, air freight logistics, delivered ex ship, road rail, transmodal,\\
& water bailage, transportive, cargon, vecturist, multiride, transfer to hospital, ...\\
\midrule
\multirow{2}{*}{building}   & tower block, nonbuilding, inbond, interior door, interiorscaper, split level,\\
& electrical wiring, seismic retrofit, house raising, sevenplex, office complex,  ...\\
\midrule
\multirow{2}{*}{river}   & mountainlike, talav, mountainside, mount sharp, river, lake albert nyanza,\\
& subapennine, khabur, transmountain, longs peak, riverling, land form, monticulus, ...\\
\midrule
\multirow{2}{*}{village}   & koprivnica, khutor, intown, b road, mini mall, oppidan, cybervillage, gaothan,\\
& lawley, shillingstone, shakespeare play, claygate, goosnargh, hamlets, northcott, ...\\
\midrule
\multirow{2}{*}{animal}   & gambian pouched rat, cattle beast, wild game, cymothoa exigua, farm animal,\\
& bestiarian, stylophora, brazilian wandering spider, western black rhinoceros, ...\\
\midrule
\multirow{2}{*}{plant}   & anthoxanthum odoratum, harpulla, calochortus amabilis, brazilian pepper tree,\\
& tree roots, cuphea, lespedeza bicolor, phoenix tree, akeake, rauli beech, nontree,...\\
\midrule
\multirow{2}{*}{album}   & studio album, lyrics, space cakes, guitar drums, song, chiodos, american life,\\
& dance pop, keys of kingdom, record deal, rock opera, songsheet, songcraft, ...\\
\midrule
\multirow{2}{*}{film}   & star actor, filmically, company men, moving pictures, stfilm, getting acquainted,\\
& sound film, photographic film, collage film, cinematology, filmize, ...\\
\midrule
\multirow{2}{*}{book}   & megabook, pilgrim's progress, neophiliac, forebook, young adult fiction, clipsheet,\\
& novels, novel, book, novelle, reading material, booklessness, e novel, ...\\
\bottomrule
\end{tabular}
\caption{Original class names and expanded class names of DBpedia.}
\label{tab:ap_class_names_db}
\end{table*}

\begin{table*}
\centering
\begin{tabular}{cc}
\toprule
\textbf{Class Name} & \textbf{Expanded Class Names}  \\ 
\midrule
\multirow{2}{*}{society, culture}   & crowd elevator, cybersociety, macroculture, intersocietal, islandness,\\
& desocialize, cultureshed, overculture, preculture, ghost skin, antisociety, ...\\
\midrule
\multirow{2}{*}{science, mathematics}   & inequality sign,ur science, odd function, common antilog, hydroscience,\\
& known quantity, find out truth, science, commutative law, aetherometry, ...\\
\midrule
\multirow{2}{*}{health}   & being well, dietetist, hale and hearty, healthcare delivery, healthful, health,\\
& country doctor, geomedical ,health centre, nutritionwise, patient contact,...\\
\midrule
\multirow{2}{*}{education, reference}   & postsecondary school, uneducation, special educator, secondary education,\\
& cross index, tertiary education, forward reference, exophora,...\\
\midrule
\multirow{2}{*}{computers, internet}   & allows null sessions, dynamic ip address, friendly url, data processor,\\
& laptops, deadlink, web diving, dictionary attacker, nt account system,  ...\\
\midrule
\multirow{2}{*}{sports}   & professional baseball, game set match, banana ball, empty bench,\\
& football, sportsman, visiting team, athletic, exhibition game, super cup, ...\\
\midrule
\multirow{2}{*}{business, finance}   & adhocratic, net operating loss, business organization, capital structure, \\
& systematic risk, manufacturers rep, web company, garmento,  ...\\
\midrule
\multirow{2}{*}{entertainment, music}   & bigophonic, good fun, entertaintment, natabhairavi, eating popcorn,\\
& allegro non troppo, semihemidemisemiquaver, musicaholic,  ...\\
\midrule
\multirow{2}{*}{family, relationships}   & mother father, enicocephalid, profamily, close friendship, salpidae,\\
& visual proximity, relations, lac scale, sexual relationship, ...\\
\midrule
\multirow{2}{*}{politics, government}   & governmentalise, ruling party, westminster system, antiindependence,\\
& leader of opposition, cryptarchy, macropolitical, antipopulist,...\\
\bottomrule
\end{tabular}
\caption{Original class names and expanded class names on AG's News.}
\label{tab:ap_class_names_yh}
\end{table*}

\begin{table*}
\centering
\begin{tabular}{cc}
\toprule
\textbf{Class Name} & \textbf{Expanded Class Names}  \\ 
\midrule
\multirow{2}{*}{bad}   & overawful, crappy, uglysome, not good, suck balls, do badder, blow chunks,\\
& shitly, godawful, sucktastic, worsts, horridsome, fucky, god awful, terrible, ...\\
\midrule
\multirow{2}{*}{good}   & correct answer, have good day, better job, clean apartment, double plus good, nice,\\
& talk with friends, goodish, supernice, like million bucks, healthy environment, ...\\
\bottomrule
\end{tabular}
\caption{Original class names and expanded class names on AG's News.}
\label{tab:ap_class_names_sent}
\end{table*}

\end{document}